\documentclass[journal, 10pt]{IEEEtran}
\IEEEoverridecommandlockouts

\makeatletter
\let\NAT@parse\undefined
\makeatother

\usepackage{amsmath}
\usepackage{amsfonts}
\usepackage{amssymb}
\usepackage{balance}
\usepackage{graphicx}
\usepackage[skip=2pt,font=small]{caption}
\usepackage{subcaption}
\usepackage[numbers, square, comma,compress]{natbib}
\usepackage{siunitx}
\usepackage{color}
\usepackage{xcolor}
\usepackage{booktabs}
\usepackage{dblfloatfix}
\usepackage{url}
\usepackage{nth}

\newcommand{\rom}[1]{(\expandafter{\romannumeral #1\relax})}
\newcommand{\mat}[1]{\begin{bmatrix}#1\end{bmatrix}}
\newcommand{\bm}[1]{\boldsymbol{#1}}
\newcommand{\mbeq}{\overset{!}{=}}
\definecolor{royalazure}{rgb}{0.0, 0.22, 0.66}
\definecolor{mayablue}{rgb}{0.45, 0.76, 0.98}

\begin{document}

\title{{CPC:} Complementary Progress Constraint\\ for Time-Optimal Quadrotor Trajectories}
\author{Philipp Foehn, Davide Scaramuzza}
\maketitle

\begin{abstract}
In many mobile robotics scenarios, such as drone racing, the goal is to generate a trajectory that passes through multiple waypoints in minimal time.
This problem is referred to as time-optimal planning.
State-of-the-art approaches either use polynomial trajectory formulations, which are suboptimal due to their smoothness, or numerical optimization, which requires waypoints to be allocated as costs or constraints to specific discrete-time nodes.
For time-optimal planning, this time-allocation is a priori unknown and renders traditional approaches incapable of producing truly time-optimal trajectories.
We introduce a novel formulation of progress bound to waypoints by a complementarity constraint.
While the progress variables indicate the completion of a waypoint, change of this progress is only allowed in local proximity to the waypoint via complementarity constraints.
This enables the simultaneous optimization of the trajectory and the time-allocation of the waypoints.
To the best of our knowledge, this is the first approach allowing for truly time-optimal trajectory planning for quadrotors and other systems.
We perform and discuss evaluations on optimality and convexity, compare to other related approaches, and qualitatively to an expert-human baseline.
\end{abstract}
\section{Introduction}
Autonomous aerial vehicles are nowadays being used for inspection, delivery, cinematography, search-and-rescue, and recently even drone racing.
The most prominent aerial system is the quadrotor, thanks to its vertical take-off and hover capabilities, its simplicity, and its versatility ranging from smooth maneuvers to extremely aggressive trajectories.
However, quadrotors have limited range dictated by their battery capacity, which in turn limits how much time can be spent on a specific task.
If the task consists of visiting multiple waypoints (delivery, inspection, drone racing \cite{Moon19jirc,Foehn20rss, Loquercio2019tro}), doing so in minimal time is often viable, and, in the context of search and rescue or drone racing, even the ultimate goal.

For simple point-mass systems, time-optimal trajectories can be computed in closed-form resulting in bang-bang acceleration trajectories, which can easily be sampled over multiple waypoints.
However, quadrotors are underactuated systems that need to rotate to adjust their acceleration direction, which always lies in the body $z$-axis.
Both the linear and rotational acceleration are controlled through the rotor thrusts, which are physically limited by the actuators.
This introduces a coupling in the achievable linear and rotational accelerations.
Therefore, time-optimal planning becomes the search for the optimal tradeoff between maximizing these accelerations.

\begin{figure}
    \centering
    \begin{subfigure}{\linewidth}
        \caption*{{\footnotesize State-of-the-Art: Waypoint Allocation using Constraints}}
        \label{fig:cpc_classic}
        \vspace{-6pt}
        \includegraphics[width=\linewidth,trim={20 60 15 0},clip]{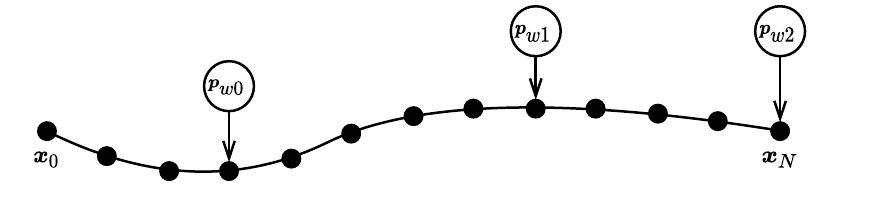}
    \end{subfigure}
    \begin{subfigure}{\linewidth}
        \caption*{{\footnotesize Our Method: Waypoint Progress Variables $\lambda$ and their Completion (Arrows)}}
        \label{fig:cpc_method}
        \vspace{-6pt}
        \includegraphics[width=\linewidth,trim={20 0 15 65},clip]{imgs/cpc.pdf}
    \end{subfigure}
    \caption{Top: state-of-the-art fixed allocation of waypoints to specific nodes.
    Bottom: our method of defining one progress variable per waypoint.
    The progress variable can switch from $1$ (incomplete) to $0$ (completed) only when in proximity of the relevant waypoint, implemented as a complementary constraint (details in Fig. \ref{fig:complementary}).}
    \label{fig:cpc}
    \vspace{-6pt}
\end{figure}

The commonly used polynomial formulations \cite{Mellinger12ijrr, Mueller13iros} for quadrotor trajectories are inherently smooth and cannot exploit the full actuator potential at every point in time, which renders them suboptimal.
This leaves planning time-discretized trajectories as the only option, which is typically solved using numerical optimization.
Unfortunately, such formulations require the allocation of waypoints as costs or constraints to specific discrete time nodes, which is a priori unknown.
We investigate this problem and provide a solution that allows to simultaneously optimize the trajectory and waypoint allocation.
Our approach formulates a progress measure for each waypoint along the trajectory, indicating completion of a waypoint (see Fig. \ref{fig:cpc}).
We then introduce a complementary progress constraint, that allows completion only in proximity to a waypoint.

There exist two common approaches in formulating the underlying state space for trajectory planning: \rom{1} polynomial representations and \rom{2} discretized state space formulations.

The first approach exploits the quadrotor's differentially-flat output states represented by smooth polynomials.
Their computation is extremely efficient and trajectories through multiple waypoints can be represented by concatenating polynomial segments, where sampling over segment times and boundary conditions provides minimal-time solutions.
However, these polynomial, minimal-time solutions are still suboptimal for quadrotors, since they
\rom{1} are smooth and cannot represent rapidly-changing states, such as bang-bang trajectories, at reasonable order,
and \rom{2} can only touch the constant boundaries in infinitesimal points, but never stay at the limit for durations different than the total segment time.
Both problems are visualized in Fig. \ref{fig:polyvsbang} and further explained in Sec. \ref{sec:preface}.

The second approach uses non-linear optimization to plan trajectories described by a time-discretized state space, where the system dynamics and input boundaries are enforced as constraints.
In contrast to the polynomial formulation, this allows the optimization to pick any input within bounds for each discrete time step.
For a time-optimal solution, the trajectory time $t_N$ is part of the optimization variables and is the sole term in the cost function.
However, if multiple waypoints must be passed, these must be allocated as constraints to specific nodes on the trajectory.
Since the fraction of overall time spent between any two waypoints is unknown, it is a priori undefined to which nodes on the trajectory such waypoint constraints should be allocated.
This problem renders traditional discretized state space formulations ineffective for time-optimal trajectory generation through multiple waypoints.

\vspace{5pt}
\noindent In this work, we answer the following research questions:
\begin{enumerate}
    \item Is it possible to revise the waypoint constraints to optimize simultaneously time-allocation, while exploiting the system dynamics and actuation limits?
    \item What are the resulting properties regarding initialization and convergence due to (non-) convex constraints?
    \item What are the characteristics of time-optimal trajectories?
\end{enumerate}

\subsection*{Contribution}
Our approach optimizes a time-discretized state and input space for minimum execution time, with the quadrotor dynamics and input bounds implemented as equality and inequality constraints, respectively.
Our contribution is the formulation of progress over the trajectory, where the completion of a waypoint is represented as a progress variable (see Fig. \ref{fig:cpc}).
We introduce a complementarity constraint, which allows the progress variable of a waypoint to switch from \textit{incomplete} to \textit{complete} only at the time nodes where the vehicle is within tolerance of the waypoint.
This allows us to encapsulate the task of reaching multiple waypoints without specifying at which time a waypoint is reached, solving the time-allocation problem.
We evaluate our approach experimentally against state-of-the-art approaches, verifying time alignment of waypoints and convergence properties, in many-waypoint scenarios, and finally comparing against human performance.

\section{Preface: Time-Optimal Quadrotor Trajectory}
\label{sec:preface}
\subsection{Point-Mass Bang-Bang}
Time-optimal trajectories encapsulate the best possible action to reach one or multiple targets in the lowest possible time.
We first investigate a point mass in $\mathcal{R}^3$ controlled by bounded acceleration $\bm{a} \in \mathcal{R}^3  ~|~ \|\bm{a}\| \leq a_{max}$, starting at rest position $\bm{p}_0$ and translating to rest position $\bm{p}_1$.
The time-optimal solution takes the form of a bang-bang trajectory over the time $t_{opt}$, accelerating with $a_{max}$ for $t_{opt}/2$, followed by decelerating with $a_{max}$ for $t_{opt}/2$.
A trajectory through multiple waypoints can be generated similarly by optimizing or sampling over the switching times and intermittent waypoint velocities.
For the general solution and applications we refer to \cite{LaValle2006book}.

\subsection{Bang-Bang Relation for Quadrotors}
\label{sec:pref_bangbang}
A quadrotor would exploit the same maximal acceleration by generating the maximum thrust.
However, due to the quadrotor's underactuation, it cannot instantaneously change the acceleration direction, but needs to rotate by applying differential thrust over its rotors.
As a result, the linear and rotational acceleration, both controlled through the limited rotor thrusts, are coupled (visualized later in Fig. \ref{fig:inputs}).
Therefore, a time-optimal trajectory is the \textit{optimal tradeoff between rotational and linear acceleration}, where the thrusts only deviate from the maximum to adjust the rotational rates.
Indeed, our experiments confirm exactly this behavior (see e.g. Fig, \ref{fig:exp_straight}).

\subsection{Sub-Optimality of Polynomial Trajectories}
\label{sec:pref_poly}
\begin{figure}[t]
    \centering
    \vspace{-3pt}
    \includegraphics[width=\linewidth]{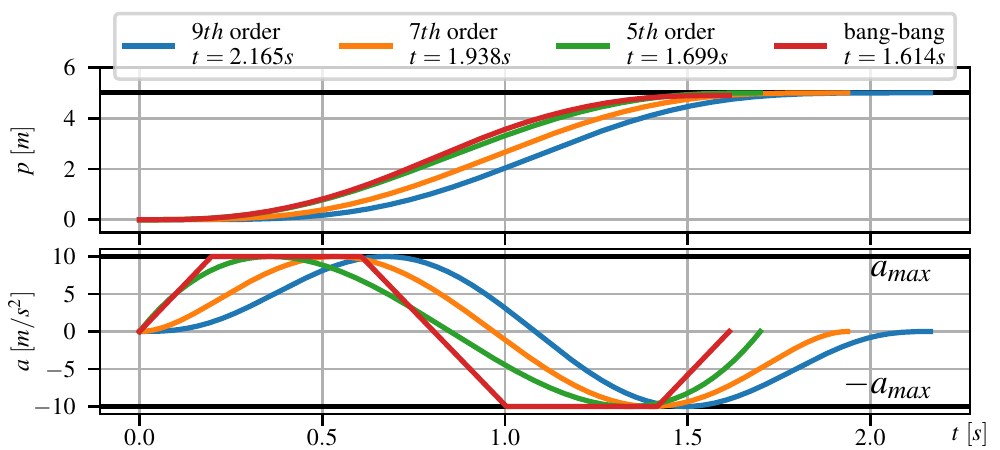}
    \caption{Multiple orders of polynomial trajectories and one bang-bang trajectory with limited slope.
    The polynomial trajectories only touch the input extrema in two points, while the bang-bang spends more time at the limit and achieves a lower overall time.}
    \label{fig:polyvsbang}
    \vspace{-12pt}
\end{figure}

The quadrotor is a differentially flat system \cite{Mellinger12ijrr} that can be described based on its four flat output states, position and yaw.
This allows to represent the evolution of the flat output states as smooth differentiable polynomials of the time $t$.
To generate such polynomials, one typically defines its boundary conditions at start and end time, and minimizes for one of the derivatives, commonly the jerk or snap (3rd/4th derivative of position, as in \cite{Mueller13iros, Mellinger11icra}).
The intention behind such trajectories is to minimize and smooth the needed body torques and, therefore, single motor thrust differences, which are dependent on the snap.
Since the polynomials are very efficient to compute (especially \cite{Mueller13iros}), trajectories through many waypoints can be generated by concatenating segments of polynomials, and minimal-time solutions can be found by optimizing or sampling over the segment times and boundary conditions.
However, these polynomials are smooth by definition, which stands in direct conflict with maximizing the acceleration at all times while simultaneously adapting the rotational rate, as explained in the previous Sec. \ref{sec:pref_bangbang}.
In fact, due to the polynomial nature of the trajectories, the boundaries of the reachable input spaces can only be touched at one or multiple points, or constantly, but not at subsegments of the trajectory.
This problem is visualized in Fig. \ref{fig:polyvsbang}, and not only applies to polynomial trajectories, but partially also to direct collocation methods for optimization \cite{Hargraves87jgcd, Posa2016icra}.

\subsection{Real-World Applications}
While the time-optimal solutions are at the boundary of the reachable input space and cannot be tracked robustly \cite{LaValle2006book}, they represent an upper-bound on the performance for a given track, intended as a baseline for other algorithms or as guidance for human pilots to improve on their skills.

\section{Related Work}

\subsection{Basic Quadrotor Control}
\label{sec:rel_basics}
The quadrotor model, its underactuated nature, and controllability have been extensively studied in \cite{Bouabdallah07iros, Mahony12ram}, with extensions to aerodynamic modelling in \cite{Faessler18ral,Tomic2020ijrr}.
Control algorithms for quadrotors are well established and reach from cascaded control \cite{Faessler17ral} to model predictive control (MPC) \cite{Bangura14ifac,Neunert16icra,Falanga18iros}.
Such MPC systems allow to track even aggressive trajectories with high reliability, thanks to their prediction and optimization over a moving horizon.
However, even those sophisticated control approaches rely on references from planners, which consider the whole time horizon of a task.

\subsection{Polynomial Trajectory Planning}
\label{sec:rel_polynomial}
The first category of such planning algorithms is based on polynomial trajectories.
The work by \cite{Mellinger11icra,Mellinger12ijrr} established the now widely used minimum-snap trajectories, minimizing the \nth{4} derivative of the position to generate smooth trajectories.

The approach by \cite{Richter2016isrr} builds on \cite{Mellinger11icra}, but extends the cost on an $n$-th order derivative by the trajectory time, effectively achieving a trade-off between lowest time and smoothness.
Exploiting polynomial formulations allows for extremely fast computation of trajectory solutions, which \cite{Mueller13iros} further exploits by limiting the approach to fifth order polynomials, which can then be solved in closed form.

In \cite{Allen16gnc} a multilevel system is proposed in the context of obstacle avoidance, first approximating the system dynamics by assuming acceleration as the bounded control input.
The simplified system is used to sample minimum-time bang-bang trajectories, which are then refined and optimized for quadrotors using polynomial representations.
By searching over boundary and intermittent conditions, or scaling the polynomials, many different problems can be solved with fast sampling based approaches.
However, as elaborated in Sec. \ref{sec:pref_poly}, the smoothness does conflict with the desired maximization of the acceleration.

\subsection{Optimization-Based Trajectory Planning}
\label{sec:rel_optimization}
Trajectory planning through numerical optimization on discretized state spaces has been widely used in the field of manipulators \cite{Ratliff09icra}, legged \cite{Posa2016icra}, and aerial robots \cite{Foehn17rss}.

To formulate trajectory planning as an optimization problem, multiple approaches exist, such as direct multiple shooting \cite{Diehl2006springer} and direct collocation \cite{Hargraves87jgcd}.
Direct collocation uses low-order splines between discrete time nodes, which can be beneficial for certain problems, but is inferior when dealing with a high number of inequality constraints.
Direct multiple shooting methods discretize the time horizon of the trajectory in $N$ steps and include both the state and inputs in the optimization variables.
The system dynamics are enforced as equality constraints over a timestep, and input limits are represented as inequality constraints.

Many approaches on numerical optimal control \cite{Neunert16icra,Falanga18iros} and trajectory generation \cite{Geisert16icra,Foehn17rss} deploy these schemes successfully for quadrotors.
They typically formulate goals and tasks as cost and/or constraints on the state and input space.
Smooth trajectories are enabled by penalizing high input deviations, similar to minimum-snap trajectories.

A simplified approach is followed by \cite{Augugliaro2012iros} in the context of multi-vehicle planning, where the optimization variables only contain the vehicles acceleration, and no rotation is represented.
Position, velocity and jerk are defined as affine functions of the acceleration, rendering this scheme a single-shooting approach with an approximative model but computationally very efficient.
Unfortunately it does only impose box-constraints on acceleration and jerk to approximate the actuation limits, which does not represent the thrust limitations well, rendering it unfit for time-optimal planning.

While these approaches allow to solve for complex tasks and maneuvers, they still rely on the user to allocate cost and constraints to specific time steps.
Normally this is not a problem, since the costs are allocated to the whole time horizon and e.g. waypoint constraints can be allocated to specific times.
However, this is no longer possible for the problem of time-optimal waypoint flight, since the exact time or fraction at which a waypoint is passed is a priori unknown.

\subsection{Time-Optimal Approaches}
\label{sec:rel_timeopt}
For simple systems, so called bang-bang or bang-singular approaches have been widely established and proven to second order sufficient conditions as in \cite{Vakhrameev1997jms,Poggiolini2006ecc,LaValle2006book}.
These approaches apply to systems with bounded input space, breaking down to a boundary value problem (e.g. switching between maximal and minimal inputs), capable of producing closed form solutions for a multitude of dynamic systems.
However, if the input bounds are or subsequent trajectory segements are not independent, these approaches become infeasible.
\cite{Hehn12ar} proposes an approach to generate time-optimal trajectories between two known states based on the aforementioned bang-bang approaches using numerical optimization of the switching times.
However, the approach is only developed for 2-dimensional maneuvers in a plane and with collective thrust and bodyrates as independent input modalities, which do not correctly represent the limited thrust inputs of a quadrotor.
In \cite{Loock13ecc} the quadrotor trajectory is represented as a convex combination of multiple paths given by analytical functions, which can additionally fulfill spatial constraints, such as obstacles.
The authors compare their approach to the one of \cite{Hehn12ar} an verify their experimental results.
However also \cite{Loock13ecc} limit their results to 2-dimensional experiments with only start and end states, without intermediate waypoints.
Compared to \cite{Loock13ecc,Hehn12ar}, our approach works on 3D trajectories through multiple waypoints while accounting for the true rotor thrust limits.

In \cite{Spedicato18tcst} a very thorough literature review and problem analysis is provided, also solving the time allocation to waypoints.
This approach uses a geometric reference path and a change of variables that puts the vehicle state space into a traverse-dynamics formulation along this path.
This allows to define progress along the track as the arc length along the geometric reference path, as well as writing cost and constraints based on the arc length and therefore independent of time.
While this approach is very elegant, it does not represent realistic actuator saturation, because it abstracts the limits as bounds on bodyrate and collective thrust, rather than single rotor thrusts, as used in our approach.
Furthermore, due to linearization characteristics of Euler angles, the orientation of the vehicle is limited, which in turn limits the space of solutions and therefore the optimality of the resulting trajectory.
Our approach allows the full orientation space with consistent linearization characteristics by using quaternions.

Finally, \cite{Ryou2020arxiv} provides a method to generate close to time-optimal segment times for polynomial trajectories using Bayesian Optimization.
The approach is based on learning Gaussian Classification models predicting feasibility of a trajectory based on multi-fidelity data from analytic models, simulation and real-world data.
While this approach puts high emphasis on real-world applicability and can even account for aerodynamic effects, it is still constrained to polynomials and requires real-world data specifically collected for the given vehicle.
Furthermore, it is an approximative method rather than the true time-optimal solution, as opposed to our method.

\section{Problem Formulation}
\subsection{General Trajectory Optimization}
The general optimization problem of finding the minimizer $\bm{x}^*$ for cost $L(\bm{x})$ in the state space $\bm{x} \in \mathcal{X}^n$ can be stated as
\begin{gather}
\bm{x}^* = \min_{x}L(\bm{x}) \\
\begin{aligned}
\text{subject to} \quad
\bm{g}(\bm{x}) &= 0 \nonumber
&\text{and} \quad
\bm{h}(\bm{x}) &\leq 0 \nonumber
\end{aligned}
\end{gather}
where $\bm{g}(\bm{x})$ and $\bm{h}(\bm{x})$ contain all equality and inequality constraints respectively.
The full state space $\bm{x}$ is used equivalent to the term \textit{optimization variables}.
The cost $L(\bm{x})$ typically contains one or multiple quadratic costs on the deviation from a reference, costs on the systems actuation inputs, or other costs describing any desired behaviors.

\subsubsection{Multiple Shooting Method}
To represent a dynamic system in the state space, the system state $\bm{x}_k$ is described at discrete times $t_k = dt \cdot k$ at $k \in (0, N)$, also called nodes, where its actuation inputs between two nodes are $\bm{u}_k$ at $t_k$ with $k \in (0, N]$.
The systems evolution is defined by the dynamics $\dot{\bm{x}} = \bm{f}_{dyn}(\bm{x}, \bm{u})$,
anchored at $\bm{x}_0=\bm{x}_{init}$,
and implemented as an equality constraint of the first order forward integration:
\begin{equation}
\bm{x}_{k+1} - \bm{x}_k - dt \cdot \bm{f}_{dyn}(\bm{x}_k, \bm{u}_k) = 0
\label{eqn:ms_forwardeuler}
\end{equation}
which is part of $\bm{g}(\bm{x})=0$ in the general formulation.
Both $\bm{x}_k$, $\bm{u}_k$ are part of the state space and can be summarized as the vehicle's dynamic states $\bm{x}_{dyn,k}$ at node $k$.

\subsubsection{Integration Scheme}
Additionally we can change the formulation to use a higher order integration scheme to suppress linearization errors for highly non-linear systems.
In this work, we deploy a 4th-order Runge-Kutta scheme:
\begin{align}
\bm{x}_{k+1} - \bm{x}_k &- dt \cdot \bm{f}_{RK4}(\bm{x}_k, \bm{u}_k) = 0 \\
\bm{f}_{RK4}(\bm{x}_k, \bm{u}_k) &=
1/6 \cdot \left( \bm{k}_1 + 2 \bm{k}_2 + 2 \bm{k}_3 + \bm{k}_4 \right)
\end{align}

\subsection{Time-Optimal Trajectory Optimization}
Optimizing for a time-optimal trajectory means that the only cost term is the overall trajectory time $L(\bm{x})=t_N$.
Therefore, $t_N$ needs to be in the optimization variables $\bm{x} = [t_N, \dots]^\intercal$, and must be positive $t_N > 0$.
The integration scheme can then be adapted to use $dt = t_N / N$.

\subsection{Passing Waypoints through Optimization}
To generate trajectories passing through a sequence of waypoints $\bm{p}_{wj}$ with $j \in [0, \dots, M]$, one would typically define a distance cost or constraint and allocate it to a specific state $\bm{x}_{dyn, k}$ at node $k$ with time $t_k$.
For cost-based formulations, quadratic distance costs are robust in terms of convergence and implemented as
\begin{equation}
L_{dist, j} = (\bm{p}_k - \bm{p}_{wj})^\intercal (\bm{p}_k - \bm{p}_{wj})    
\end{equation}
where $\bm{p}_k$, part of $\bm{x}$, is the position state at a user defined time $t_k$.
However, such a cost-based formulation is only a soft requirement and if summed with other cost terms does not imply that the waypoint is actually passed within a certain tolerance.
To guarantee passing within a tolerance, constraint-based formulations can be used, such as
\begin{equation}
(\bm{p}_k - \bm{p}_{wj})^\intercal (\bm{p}_k - \bm{p}_{wj}) \leq \tau_j^2    
\end{equation}
which in the general problem is part of $\bm{h}(\bm{x}) \leq 0$, and requires the trajectory to pass by waypoint $j$ at position $\bm{p}_{wj}$ within tolerance $\tau_j$ at time $t_k$.

\subsection{\textbf{The Problem} of Time Allocation and Optimization}
Adding waypoints to a trajectory means that their costs or constraints need to be allocated to specific nodes (at times $t_k$) on the trajectory.
This fixes the time at which a waypoint is passed, or even if we optimize over the total time $t_N$, it still fixes the fractional time $t_k = t_N / N \cdot k$.
However, it is not always possible to know a priori how much time or what fraction of the total time is spent between two waypoints.
Therefore, it is not possible to allocate waypoint costs or constraints to nodes at specific times.

One approach is to spread the cost of reaching the waypoint over multiple nodes, which is done in \cite{Neunert16icra} with an exponential weight spread.
Even though this allows to shift the time of passing a waypoint, the width and mean of the weight spread are still user-defined, suboptimal in most scenarios, and does not allow to freely shift the time at which a waypoint is passed.

There are two possible ways to solve this: \rom{1} change the time between fixed nodes, which implies changing $dt$ for each timestep by adding all $dt_k$ to the optimization variables; or \rom{2} change the node $k$ to which a waypoint is allocated.
The first option allows varying timeteps $dt_k$, which negatively impacts the linearization quality and makes it inconsistent over the trajectory.
This allows infeasible solutions, where all but a few $dt_k$ are set to zero, rendering the integration scheme ineffective, and leading to a violation of the real quadrotor dynamics.
The second option requires a formulation that allows the optimization to change the node $k$ to which a waypoint is allocated.
Since this tackles the fundamental underlying allocation problem, we base our approach on this second option.

\subsection{Problem Formulation Summary}
\textit{
In conclusion, the goal of this work is to find an optimization problem formulation which
\rom{1} satisfies the system dynamics and input limitations,
\rom{2} minimizes total trajectory time,
\rom{3} passes by multiple waypoints in sequence,
and
\rom{4} finds the optimal time at which to pass each waypoint.
}
\clearpage
\section{Approach:\\Complementary Progress Constraints}
\label{sec:approach}
Our approach consists of two main steps: \rom{1} adding a measure of progress throughout the track, and \rom{2} adding a measure of how this progress changes.
In the following sections we explain how these two steps can be formulated and added to an numerical optimization problem.

\subsection{Progress Measure Variables}
To describe the progress throughout a track we want a measure that fulfills the following requirements: \rom{1} it starts at a defined value, \rom{2} it must reach a different value by the end of the trajectory, and \rom{3} it can only change when a waypoint is passed within a certain tolerance.
To achieve this, let the vector $\boldsymbol{\lambda}_k \in \mathcal{R}^M$ define the progress variables $\lambda_k^j$ at timestep $t_k$ for all $M$ waypoints indexed by $j$.
All progress variables start at 1 as in $\boldsymbol{\lambda}_0 = \bm{1}$ and must reach 0 at the end of the trajectory as in $\boldsymbol{\lambda}_N = \bm{0}$.
The progress variables $\boldsymbol{\lambda}$ are chained together and their evolution is defined by
\begin{equation}
\boldsymbol{\lambda}_{k+1} = \boldsymbol{\lambda}_k - \boldsymbol{\mu}_k
\end{equation}
where the vector $\boldsymbol{\mu}_k \in \mathcal{R}^M$ indicates the progress change at every timestep.
Note that the progress can only be positive, therefore $\mu_k^j \geq 0$.
Both $\boldsymbol{\lambda}_k$ and $\boldsymbol{\mu}_k$ for every timestep are part of the optimization variables $\bm{x}$, which replicates the multiple shooting scheme for the progress variables.

To define when and how the progress variables can change, we now imply constraints on $\boldsymbol{\mu}_k$, in it's general form as
\begin{align}
\bm{\epsilon}^-_k \leq \bm{f}_{prog}(\bm{x}_k, \bm{\mu}_k) \leq \bm{\epsilon}^+_k
\end{align}
where $\bm{\epsilon}^-$, $\bm{\epsilon}^+$ can form equality or inequality constraints.

Finally, to ensure that the waypoints are passed in the given sequence, we enforce subsequent progress variables to be bigger than their prequel at each timestep by
\begin{equation}
\mu_k^j \leq \mu_k^{j+1} \quad \forall \quad k \in [0, N], j \in [0, M).
\end{equation}

\subsection{Complementary Progress Constraints}
In the context of waypoint following, the goal is to allow $\boldsymbol{\mu}_k$ to only be non-zero at the time of passing a waypoint.
Therefore, $\bm{f}_{prog}$ and $\bm{\epsilon}^- = \bm{\epsilon}^+ = 0$ are chosen to represent a \textit{complementarity constraint} \cite{Posa2016icra}, as
\begin{equation}
\bm{f}_{prog}(\bm{x}_k, \bm{\mu}_k) = \left[ \mu_k^j \cdot \|\bm{p}_k - \bm{p}_{wj}\|_2^2 \right]_{j \in [0, M)} \mbeq 0
\label{eqn:compconst}
\end{equation}
which can be interpreted as a mathematical \textit{OR} function, since either $\mu_k^j$ or $\|\bm{p}_k - \bm{p}_{wj}\|$ must be $0$.
Intuitively, the two elements \textit{complement} each other.

\subsection{Tolerance Relaxation}
With \eqref{eqn:compconst} the trajectory is forced to pass \textit{exactly} through a waypoint.
Not only is this impractical, since often a certain tolerance is admitted or even wanted, but it also negatively impacts the convergence behavior and time-optimality, since the system dynamics are discretized and one of the discrete timesteps must coincide with the waypoint.
Therefore, it is desirable to relax a waypoint constraint by a certain tolerance which is achieved by extending \eqref{eqn:compconst} to
\begin{gather}
\bm{f}_{prog}(\bm{x}_k, \bm{\mu}_k) = \left[ \mu_k^j \cdot \left( \|\bm{p}_k - \bm{p}_{wj}\|_2^2 - \nu_k^j\right) \right]_{j \in [0, M)} \overset{!}{=} 0 \nonumber \\
\text{subject to} \quad 0 \leq \nu_k^j \leq d_{tol}^2 \label{eqn:compconsttol}
\end{gather}
where $\nu_k^j$ is a slack variable to allow the distance to the waypoint to be relaxed to zero when it is smaller than $d_{tol}$, the maximum distance tolerance.
This now enforces that the progress variables can not change, except for the timesteps at which the system is within tolerance to the waypoint.

\begin{figure}[t]
    \centering
    \includegraphics[width=0.9\linewidth, trim={120 0 120 20}, clip]{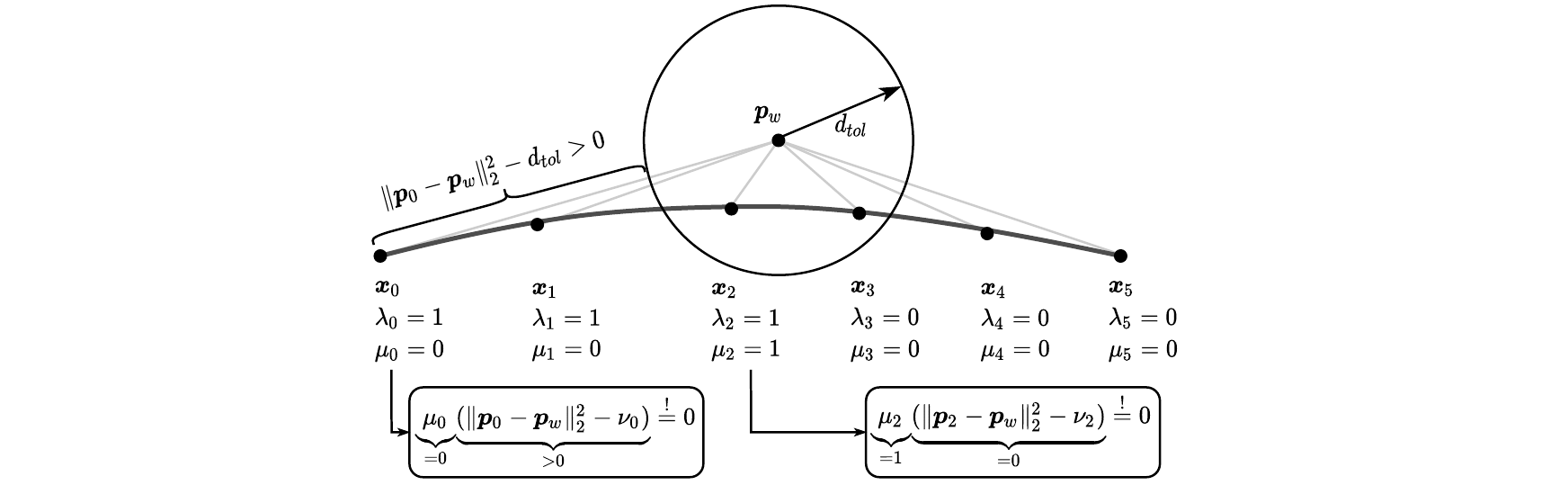}
    \caption{Illustration of the Complementary Progress Constraint.
    $\mu$ can only be non-zero if the distance to the waypoint $\bm{p}_w$ is less than the tolerance $d_{tol}$.
    This is not the case for $\bm{x}_0$, but for $\bm{x}_1$, and allowing the progress variable to switch to $0$ (complete).}
    \label{fig:complementary}
\end{figure}

\section{Full Problem Formulation}
The full space of optimization variables $\bm{x}$ consists of the overall time and all variables assigned to nodes $k$ as $\bm{x}_k$.
All nodes $k$ include the robot's dynamic state $\bm{x}_{dyn,k}$, its inputs $\bm{u}_k$, and all progress variables, therefore: 
\begin{align}
\bm{x} &= [t_N, \bm{x}_0, \dots, \bm{x}_N ] \\
\text{where} & \nonumber \\
\bm{x}_k &= 
\begin{cases}
[\bm{x}_{dyn, k}, \bm{u}_k, \boldsymbol{\lambda}_k, \boldsymbol{\mu}_k, \boldsymbol{\nu}_k] &  \text{for } k \in [0, N) \\
[\bm{x}_{dyn, N}, \boldsymbol{\lambda}_N] & \text{for } k = N.
\end{cases}
\end{align}
\\
Based on this representation, we write the full problem as
\begin{equation}
\bm{x}^* = \min_{\bm{x}} t_N
\end{equation}
subject to the system dynamics and initial constraint
\begin{gather}
\bm{x}_{k+1} - \bm{x}_k - dt \cdot \bm{f}_{RK4}(\bm{x}_k, \bm{u}_k) = 0 \\
\bm{x}_0 = \bm{x}_{init},
\end{gather}
the input constraints
\begin{align}
\bm{u}_{min} - \bm{u}_k &\leq 0 &
\bm{u}_k - \bm{u}_{max} &\leq 0,
\end{align}
the progress evolution, boundary, and sequence constraints 
\begin{gather}
\bm{\lambda}_{k+1} - \bm{\lambda}_k + \bm{\mu}_k = 0 \\
\begin{aligned}
\bm{\lambda}_0 -1 &= 0 &
\bm{\lambda}_N &= 0 
\end{aligned}\\
\lambda_k^j - \lambda_k^{j+1} \leq 0 \quad \forall j \in [0, m),
\end{gather}
and the complementarity constraint with tolerance
\begin{gather}
\mu_k^j \cdot \left( \| \bm{p}_k - \bm{p}_{wj} \|_2^2 - \nu_k^j \right) = 0 \\
\begin{aligned}
-\nu_k^j &\leq 0 & \quad
\nu_k^j - d_{tol}^2 &\leq 0.
\end{aligned}
\end{gather}

\section{Application to Quadrotors}
\label{sec:quadrotor}
\begin{figure}[t]
\begin{subfigure}{0.45\linewidth}
  \centering
  \caption{Acceleration Space}
  \label{fig:inputs_acc}
  \vspace{-5pt}
  \includegraphics[width=\linewidth]{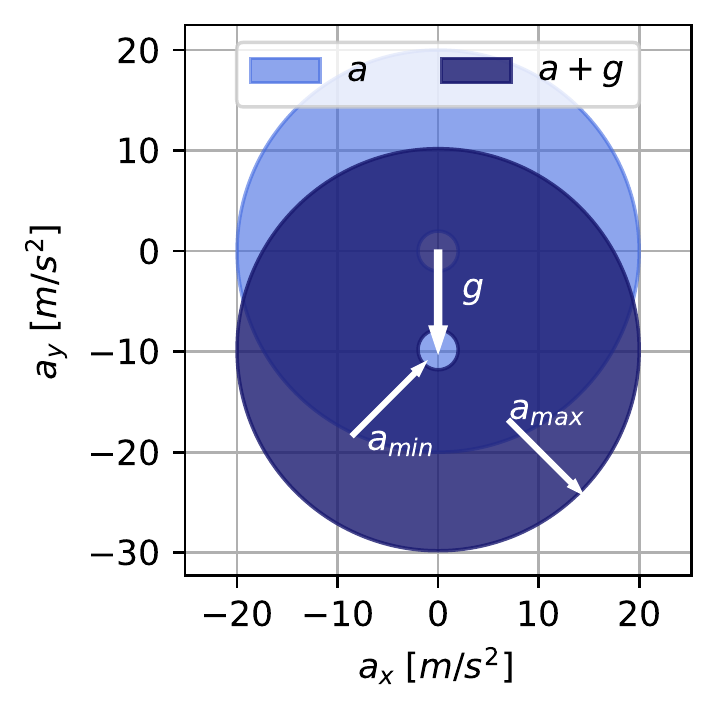}
\end{subfigure}
\hfill
\begin{subfigure}{0.45\linewidth}
  \centering
  \caption{Thrust and Torque Space}
  \label{fig:inputs_tt}
  \vspace{-5pt}
  \includegraphics[width=\linewidth]{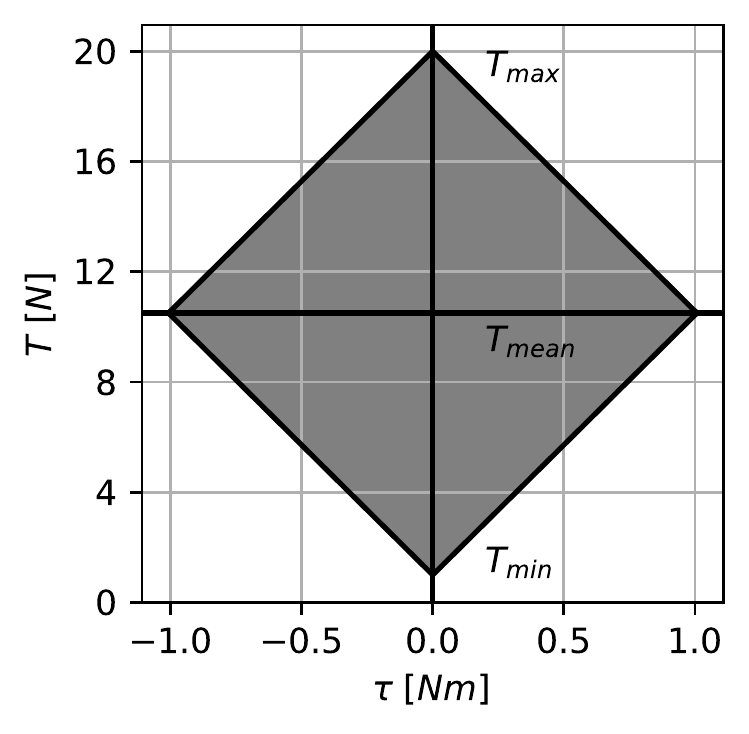}
\end{subfigure}
\caption{Acceleration- (Fig.\ref{fig:inputs_acc}) and thrust/torque-space (Fig. \ref{fig:inputs_tt}) of the STD quadrotor configuration (see Tab. \ref{tab:quads}).
Note that the acceleration space in Fig. \ref{fig:inputs_acc} is non-convex due to $T_{min} > 0$ and the thrust and torque limits are dependent on each other in \ref{fig:inputs_tt}.}
\label{fig:inputs}
\end{figure}
To apply our time-optimal planning method to a quadrotor, we first define its state space, input modality, and dynamics.

\subsection{Quadrotor Dynamics}
The quadrotor's state space is described between the inertial frame $I$ and body frame $B$, as $\bm{x} = [\bm{p}_{IB}, \bm{q}_{IB}, \bm{v}_{IB}, \bm{w}_{B}]^\top$ corresponding to position $\bm{p}_{IB}\in \mathbb{R}^3$, unit quaternion rotation $\bm{q}_{IB} \in \mathbb{R}^4$ given $\| \bm{q}_{IB} \|=1$, velocity $\bm{v}_{IB} \in \mathbb{R}^3$, and bodyrate $\bm{\omega}_{B} \in \mathbb{R}^3$.
The input modality is on the level of collective thrust $\bm{T}_B = \mat{0 & 0 & T_{Bz}}^\intercal$ and body torque $\bm{\tau}_B$.
From here on we drop the frame indices since they are consistent throughout the description.
The dynamic equations are
\begin{align}
\dot{\bm{p}} &= \bm{v} &
\dot{\bm{q}} &= \frac{1}{2} \bm{\Lambda} (\bm{q}) \mat{0 \\ \boldsymbol{\omega}} \\ 
\dot{\mathbf{v}} &= \mathbf{g} + \frac{1}{m} \mathbf{R}(\mathbf{q}) \mathbf{T} &
\dot{\boldsymbol{\omega}} &= \mathbf{J}^{-1} \left( \boldsymbol{\tau} - \boldsymbol{\omega} \times \mathbf{J} \boldsymbol{\omega} \right)
\end{align}
where $\boldsymbol{\Lambda}$ represents a quaternion multiplication, $\bm{R}(\bm{q})$ the quaternion rotation, $m$ the quadrotor's mass, and $\bm{J}$ its inertia.

\subsection{Quadrotor Inputs}
The input space given by $\bm{T}$ and $\bm{\tau}$ is further decomposed into the single rotor thrusts $\mathbf{u} = [T_1, T_2, T_3, T_4]$. where $T_i$ is the thrust at rotor $i \in \{ 1, 2, 3, 4 \}$.
\begin{align}
\bm{T} &= \mat{0 \\ 0 \\ \sum T_i} &
\text{and }
\boldsymbol{\tau} &=
\mat{l/\sqrt{2} (T_1 + T_2 - T_3 - T_4) \\
l/\sqrt{2} (- T_1 + T_2 + T_3 - T_4) \\
c_\tau (T_1 - T_2 + T_3 - T_4)}
\end{align}
with with the quadrotor's arm length $l$ and the rotor's torque constant $c_\tau$.
The quadrotor's actuators limit the applicable thrust for each rotor, effectively constraining $T_i$ as 
\begin{equation}
0 \leq T_{min} \leq T_i \leq T_{max}
\end{equation}
where $T_{min}$, $T_{max}$ are positive for typical quadrotor setups.
In Fig. \ref{fig:inputs} we visualize the acceleration space and the thrust torque space of a quadrotor in the $xz$-plane.
Note that the acceleration space in Fig. \ref{fig:inputs_acc} is not convex due to $T{min} > 0$ for the depicted model parameters from the STD configuration of Tab. \ref{tab:quads}.
The torque space is visualized in Fig. \ref{fig:inputs_tt}, where the coupling between the achievable thrust and torque is visible.

\subsection{Approximative Linear Aerodynamic Drag}
Finally, we extend the quadrotor's dynamics to include a linear drag model, to approximate the most dominant aerodynamic effects with drag coefficient $c_D$ by
\begin{equation}
\dot{\bm{v}} = \bm{g} + \frac{1}{m} \bm{R}(\bm{q}) \bm{T} - c_D \cdot \bm{v}.
\end{equation}
We approximate $c_D = \sqrt{\left( 4 \cdot T_{max} / m \right)^2 - g^2} / v_{max}$ to cancel the full thrust in horizontal steady-state flight at $v_{max}$. 

\section{Experimental Evaluation}
\label{sec:experiments}
To demonstrate the capabilities and applicability of our method, we test it on a series of experiments.
We first evaluate a simple point-to-point scenario and compare to \cite{Hehn12ar, Loock13ecc} in Sec. \ref{sec:exp_p2p}.
Next, we investigate the time alignment for multiple waypoints in Sec. \ref{sec:exp_time}, followed by an experimental investigation of the convergence characteristics on short tracks in terms of initialization in Sec. \ref{sec:exp_initconv} and (non-) convexity in Sec \ref{sec:exp_nonconv}.
Then we demonstrate applicability to longer tracks with $\geq 10$ waypoints (Sec. \ref{sec:exp_slalom}, \ref{sec:exp_airsim}).
Finally, we compare time-optimal trajectories to human trajectories flown in simulation (Sec. \ref{sec:exp_humansim} and real-world (Sec. \ref{sec:exp_humanreal}). 

All evaluations are performed using CasADI \cite{Andersson18} with IPOPT \cite{Waechter06jmp} as solver backend.
We use multiple different quadrotor configurations, as listed in Tab. \ref{tab:quads}.
The first configuration (RQ) represents a typical race quadrotor, the second one (MS) is parameterized after the MicroSoft AirSim \cite{Shah17fsr} SimpleFlight quadrotor, and the third  (SIM) resembles the drone used in simulation flown by a human pilot.
The last standard configuration (STD) resembles the one used in \cite{Hehn12ar, Loock13ecc}.

\begin{table}[t]
    \centering
    \setlength{\tabcolsep}{3pt}
    \caption{Quadrotor Configurations}
    \label{tab:quads}
    {\footnotesize
    \begin{tabular}{l|c|c|c|c}
        \toprule
        Property & RQ & MS & SIM & STD \\
        \midrule
        $m$ $[\si{\kilo\gram}]$ & $0.76$ & $1.0$ & $3.2$ & $1.0$ \\
        $l$ $[\si{\meter}]$ & $0.17$ & $0.23$ & $0.232$  & $0.15$ \\
        $diag(J)$ $[\si{\gram\meter^2}]$ & $[3, 3, 5]$ & $[10, 10 ,20]$ & $[50, 23, 67]$ & $[5, 5, 10]$ \\
        $T_{min}$ $[\si{\newton}]$ & $0.0$ & $0.0$ & $0.5$ & $0.25$ \\
        $T_{max}$ $[\si{\newton}]$ & $16.0$ & $4.179$ & $12$ & $5.0$ \\
        $c_\tau$ $[1]$ & $0.01$ & $0.0133$ & $0.0133$ & $0.01$ \\
        $\omega_{max}$ $[\si{\radian\per\second}]$ & $15$ & $10$ & $3$ & $10$ \\
        $v_{max}$ $[\si{\meter\per\second}]$ & $42$ & $19$ & $20$ & $-$ \\
        \bottomrule
    \end{tabular}
    }
    \vspace{-8pt}
\end{table}

\subsection*{Initialization Setup}
If not stated differently, the optimization is initialized with identity orientation, zero bodyrates, $\SI{1}{\meter\per\second}$ velocity, linearly interpolated position between the waypoints, and hover thrusts.
The total time is set as the distance through all waypoints divided by the velocity guess.
The node of passing a waypoint (respectively where the progress variables $\bm{\lambda}$ switch to zero) is initialized as equally distributed, i.e. for waypoint $j$ the passing node is $k_j = N \cdot j / M$.
We define the total number of nodes $N$ based on the number of nodes per waypoint $N_w$ so that $N = M N_w$.
We typically chose roughly $N_w \in (50, 100)$ nodes per waypoint, to get a good linearization depending on the overall length, complexity, and time of the trajectory.
Note that high numbers of $N_w \gg 100$ help with convergence and achieved stability of the underlying optimization algorithm, but on longer tracks can cause very long computation times.

\begin{figure*}[b]
    \centering
    \begin{subfigure}{0.48\linewidth}
        \caption{Regular Waypoint Distribution $t_N = \SI{2.340}{\second}$}
        \label{fig:exp_straight_regular}
        \begin{subfigure}{\linewidth}
        \centering
        \caption*{$xy$-position}
        \vspace{-5pt}
        \includegraphics[width=\textwidth,trim={0 190 0 10},clip]{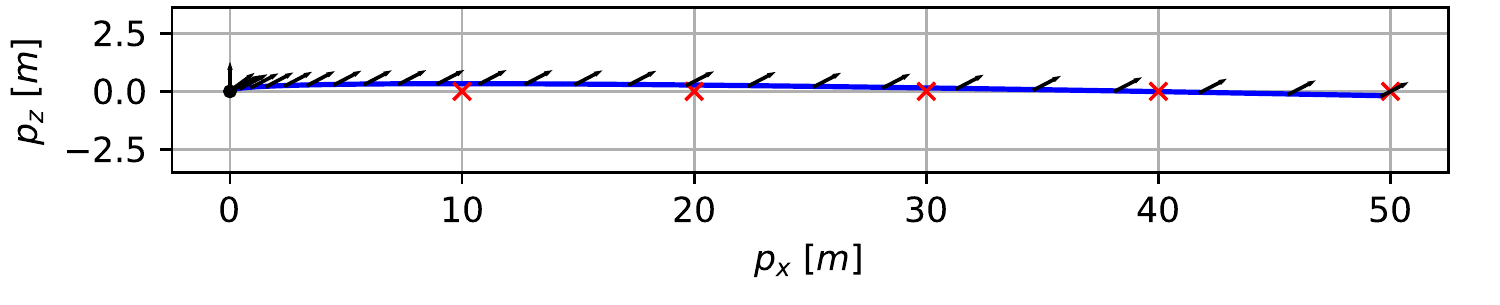}
        \end{subfigure}
        \begin{subfigure}{\linewidth}
        \centering
        \caption*{Inputs}
        \vspace{-5pt}
        \includegraphics[width=\textwidth,trim={0 100 0 105},clip]{imgs/straight.pdf}
        \end{subfigure}
        \begin{subfigure}{\linewidth}
        \centering
        \caption*{Progress}
        \vspace{-5pt}
        \includegraphics[width=\textwidth,trim={0 10 0 195},clip]{imgs/straight.pdf}
        \end{subfigure}
    \end{subfigure}
    \begin{subfigure}{0.48\linewidth}
        \centering
        \caption{Irregular Waypoint Distribution $t_N = \SI{2.340}{\second}$}
        \label{fig:exp_straight_irregular}
        \begin{subfigure}{\linewidth}
        \centering
        \caption*{$xy$-position}
        \vspace{-5pt}
        \includegraphics[width=\textwidth,trim={0 190 0 10},clip]{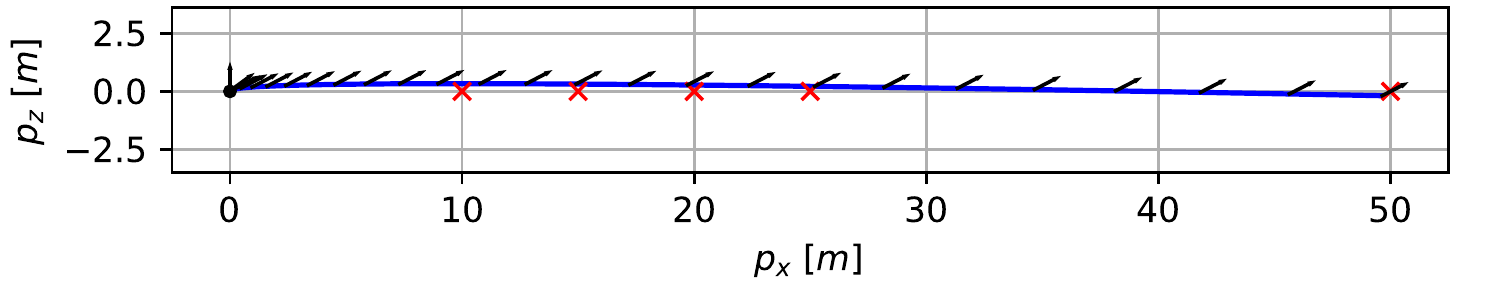}
        \end{subfigure}
        \begin{subfigure}{\linewidth}
        \centering
        \caption*{Inputs}
        \vspace{-5pt}
        \includegraphics[width=\textwidth,trim={0 100 0 105},clip]{imgs/straight_squeezed.pdf}
        \end{subfigure}
        \begin{subfigure}{\linewidth}
        \centering
        \caption*{Progress}
        \vspace{-5pt}
        \includegraphics[width=\textwidth,trim={0 10 0 195},clip]{imgs/straight_squeezed.pdf}
        \end{subfigure}
    \end{subfigure}
    \vspace{4pt}
    \caption{
    A trajectory along waypoints distributed on a line over $\SI{50}{\meter}$, flown in $\SI{2.430}{\second}$.
    In Fig. \ref{fig:exp_straight_regular} the waypoints are equally distributed over the total distance, while in Fig. \ref{fig:exp_straight_irregular} the first 4 out of 5 waypoints are located within the first half of the total distance.
    The bottom plot depicts the progress variables as initialized (dashed $--$) and as in the final solution (solid $-$).
    Note that both settings converge to the exact same solution, while the time of passing the waypoints was significantly adjusted from the initialization, and is different between the settings.}
    \label{fig:exp_straight}
\end{figure*}

\newpage
\subsection{Time-Optimal Hover-to-Hover Trajectories}
\label{sec:exp_p2p}
We first evaluate trajectory generation between two known position states in hover, one at the origin, and one at $p_x = [3, 6, 9, 12, 15]\si{\meter}$, as in \cite{Hehn12ar,Loock13ecc}.
Additionally to the problem setup explained in Sec. \ref{sec:approach}, we add constraints to the end state to be in hover, i.e. $\bm{v}_N=\bm{0}$ and $\bm{q}=\mat{1 & 0 & 0 & 0}$.
We use $N=N_w=50$ nodes and a tolerance of $d_{tol}=10^{-3}$.
Different from \cite{Hehn12ar,Loock13ecc}, we compute the solution in full 3D space, which however does not matter for this experiment, since the optimal trajectory stays withing the $y$-plane.
We defined the model properties so that it meets the maximal and minimal acceleration $[a_{min}, a_{max}] = [1, 20]\si{\meter\per\square\second}$ and maximal bodyrate $\omega_{max}=\SI{10}{\radian\per\second}$ as in \cite{Hehn12ar,Loock13ecc}, given by our STD configuration (see Tab. \ref{tab:quads}).

\begin{figure}[t]
    \centering
    \begin{subfigure}{\linewidth}
        \caption{Hover-to-Hover Trajectories}
        \label{fig:exp_p2p_pos}
        \vspace{-6pt}
        \includegraphics[width=0.95\linewidth]{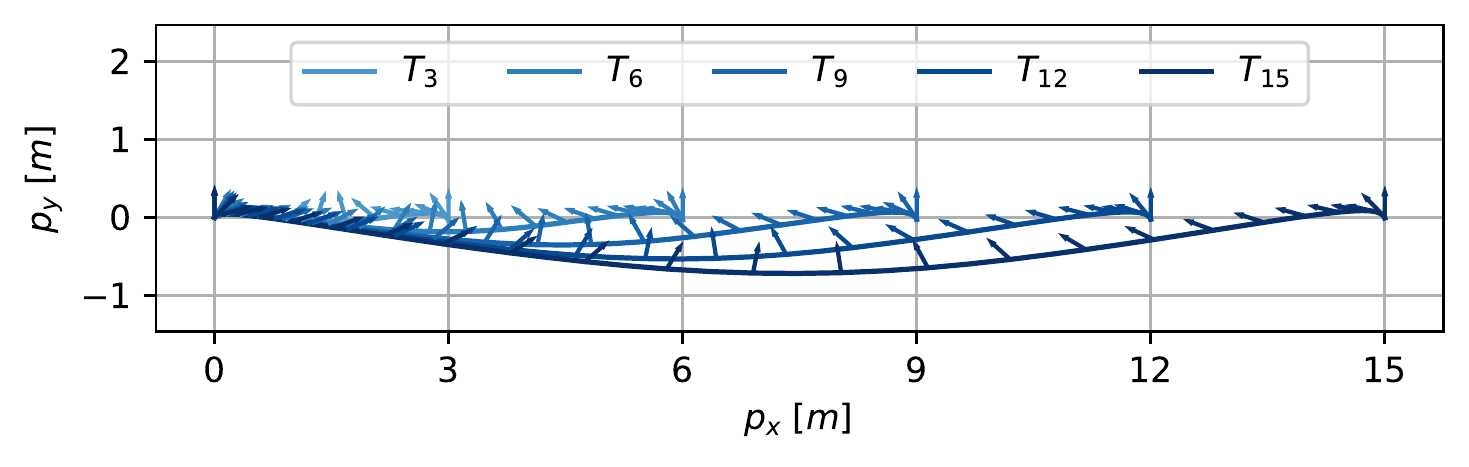}
    \end{subfigure}
    \begin{subfigure}{\linewidth}
        \caption{Velocity Profile}
        \label{fig:exp_p2p_vel}
        \vspace{-6pt}
        \includegraphics[width=0.95\linewidth]{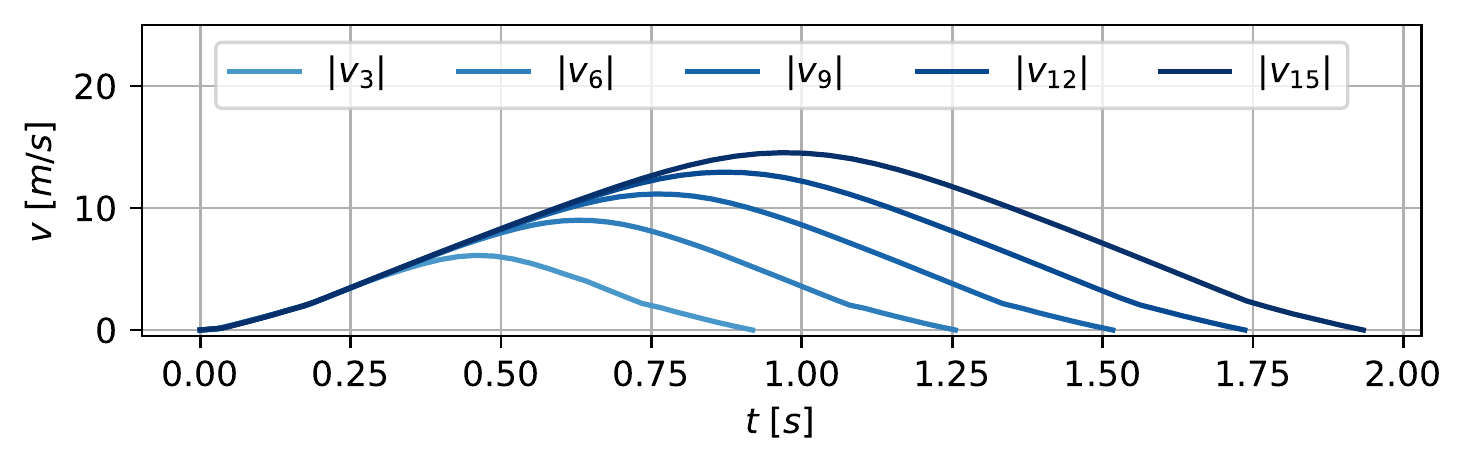}
    \end{subfigure}
    \caption{Time-optimal hover-to-hover trajectories between states spaced $p_x=[3, 6, 9, 12, 15]\si{\meter}$ apart as in \cite{Hehn12ar,Loock13ecc}.
    The top Fig. \ref{fig:exp_p2p_pos} depicts the position on the $xy$-plane, while the lower Fig. \ref{fig:exp_p2p_vel} depicts the velocity profile.}
    \label{fig:exp_p2p}
    \vspace{-12pt}
\end{figure}

The solutions are depicted in the $xy$-plane in Fig. \ref{fig:exp_p2p} and the timings are stated and compared to \cite{Hehn12ar,Loock13ecc} in Tab. \ref{tab:exp_p2p}.
Note that our approach is $2.00\%$ slower than \cite{Hehn12ar} and $2.68\%$ slower than \cite{Loock13ecc} because, differently from those works, it also models the full rotational dynamics and accounts for realistic actuation limits.
Furthermore, \cite{Hehn12ar, Loock13ecc} do not allow to compute trajectories through multiple waypoints, while our method does and simultaneously solves the time-allocation, as evaluated in the next section.

\begin{table}[t]
    \centering
    \begin{tabular}{lccc}
        \toprule
        $p_x$ & Time of \cite{Hehn12ar} & Time of \cite{Loock13ecc} & Time of ours \\
        \midrule
        $\SI{3}{\meter}$ & $\SI{0.898}{\second}$ & $\SI{0.890}{\second}$ & $\SI{0.918}{\second}$ \\
        $\SI{6}{\meter}$ & $\SI{1.231}{\second}$ & $\SI{1.223}{\second}$ & $\SI{1.255}{\second}$ \\
        $\SI{9}{\meter}$ & $\SI{1.488}{\second}$ & $\SI{1.478}{\second}$ & $\SI{1.517}{\second}$ \\
        $\SI{12}{\meter}$  & $\SI{1.705}{\second}$ & $\SI{1.694}{\second}$ & $\SI{1.736}{\second}$ \\
        $\SI{15}{\meter}$  & $\SI{1.895}{\second}$ & $\SI{1.885}{\second}$ & $\SI{1.933}{\second}$ \\
        \bottomrule
    \end{tabular}
    \caption{Comparison of the resulting timings between our approach and \cite{Hehn12ar,Loock13ecc}. Note that our approach is $2.00\%$ slower than \cite{Hehn12ar} and $2.68\%$ slower than \cite{Loock13ecc}, because it accounts for rotation dynamics and true actuator limits.}
    \label{tab:exp_p2p}
\end{table}

\subsection{Optimal Time Allocation on Multiple Waypoints}
\label{sec:exp_time}
In this experiment, we define a straight track between origin and $p_x = \SI{50}{\meter}$ trough multiple waypoints.
The goal is to show how our method can choose the optimal time at which a waypoint is passed, and we expect to see both setups to converge to the same solution.
Therefore, we test two different distributions of the waypoints $\bm{p}_{wj}$ over the straight track; specifically, we define a regular ($p_{x,reg} = [1, 20, 30, 40, 50]\si{\meter}$), and an irregular ($p_{x,ireg} = [10, 15, 20, 25, 50]\si{\meter}$) distribution.
We chose $N=125$ and a tolerance of $d_{tol}=\SI{0.4}{\meter}$, with the STD quadrotor (see Table \ref{tab:quads}).

As expected, both setups converge to the same solution of $t_N = \SI{2.430}{\second}$, depicted in Fig. \ref{fig:exp_straight}, with equal state and input trajectories, despite the different waypoint distribution.
Since the waypoints are located at different intervals, we can observe a different distribution of the progress variables in Fig. \ref{fig:exp_straight}, while the trajectory time, dynamic states, and inputs stay the same.

\begin{figure*}[t]
    \begin{subfigure}{0.49\linewidth}
    \centering
        \caption{Initialized passing through both waypoints.\quad $t_N = \SI{3.600}{\second}$}
        \label{fig:exp_hairpin_pos}
        \vspace{-3pt}
        \includegraphics[width=\textwidth,trim={0 0 0 25},clip]{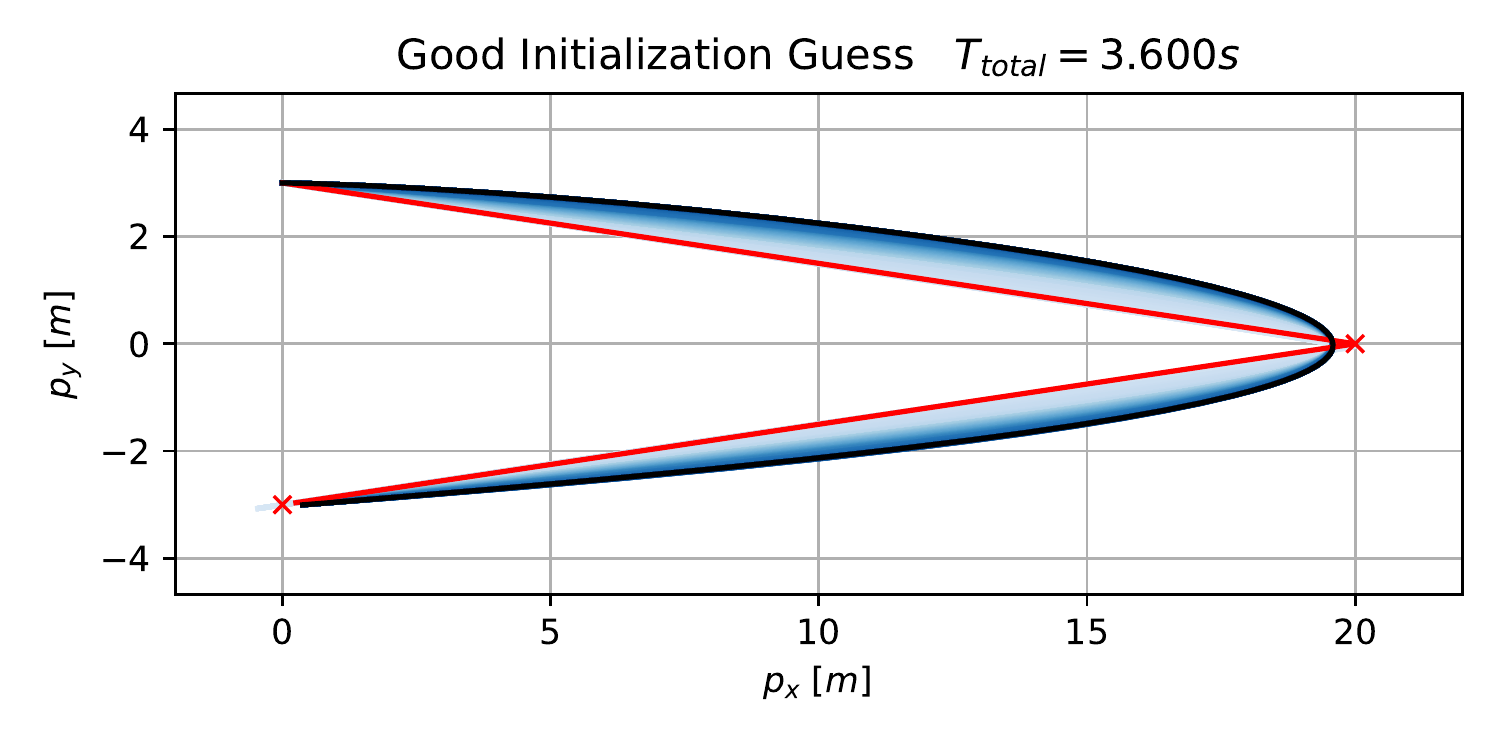}
    \end{subfigure}
    \hfill
    \begin{subfigure}{0.49\linewidth}
        \centering
        \caption{Initialized passing through only one waypoints.\quad $t_N = \SI{3.600}{\second}$}
        \label{fig:exp_hairpin_init_pos}
        \vspace{-3pt}
        \includegraphics[width=\textwidth,trim={0 0 0 25},clip]{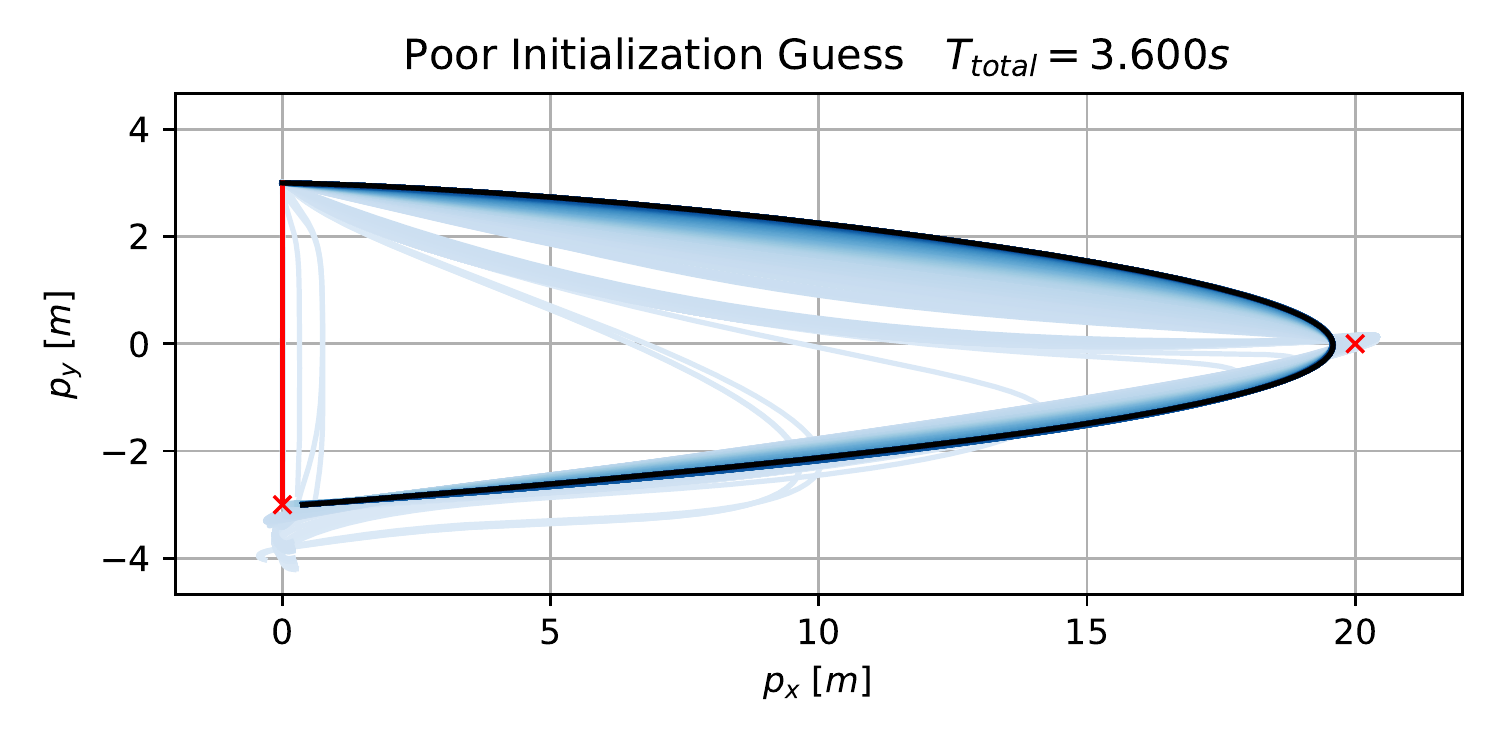}
    \end{subfigure}
    \vspace{-3pt}
    \caption{Convergence in an open hairpin turn from two different initializations in red ({\color{red} ---}) over the iterations in light blue ({\color{mayablue} ---}), to the final solution in dark blue({\color{royalazure} ---}).
    The trajectory starts at the top left and passes through the two waypoints ({\color{red} $\times$}).
    While the good initialization of Fig. \ref{fig:exp_hairpin_pos} needs $216$ iterations, the poor guess in Fig. \ref{fig:exp_hairpin_init_pos} needs $303$ iterations, but both converge to exactly the same solution with $t_N = \SI{3.600}{\second}$.}
    \label{fig:exp_hairpin}
\end{figure*}
\begin{figure*}[b]
    \begin{subfigure}{0.5\linewidth}
        \centering
        \caption{Non-Convex Acceleration Space}
        \label{fig:exp_vertpin_std}
        \includegraphics[width=\linewidth,trim={0 0 0 21},clip]{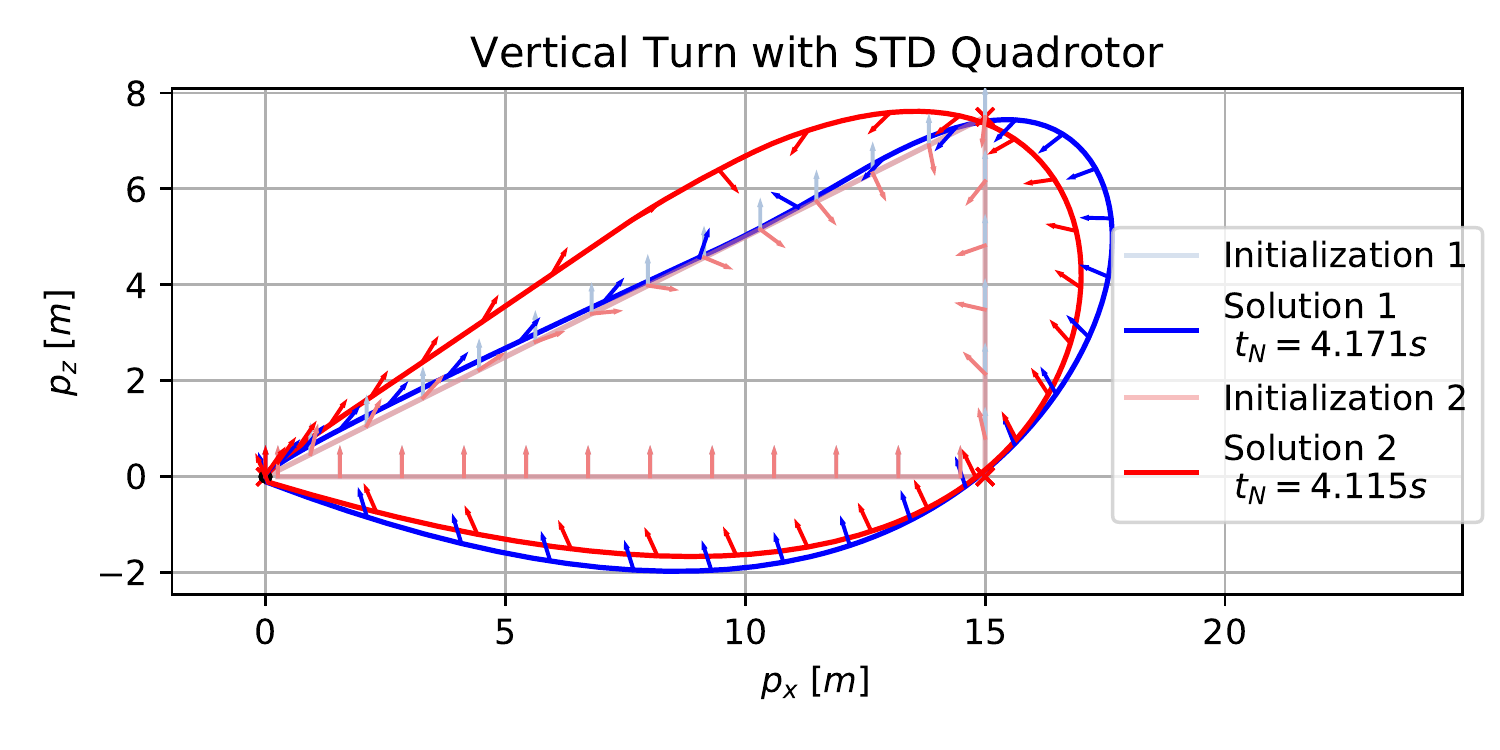}
    \end{subfigure}
    \hfill
    \begin{subfigure}{0.5\linewidth}
        \centering
        \caption{Convex Acceleration Space.}
        \label{fig:exp_vertpin_rq}
        \includegraphics[width=\linewidth,trim={0 0 0 21},clip]{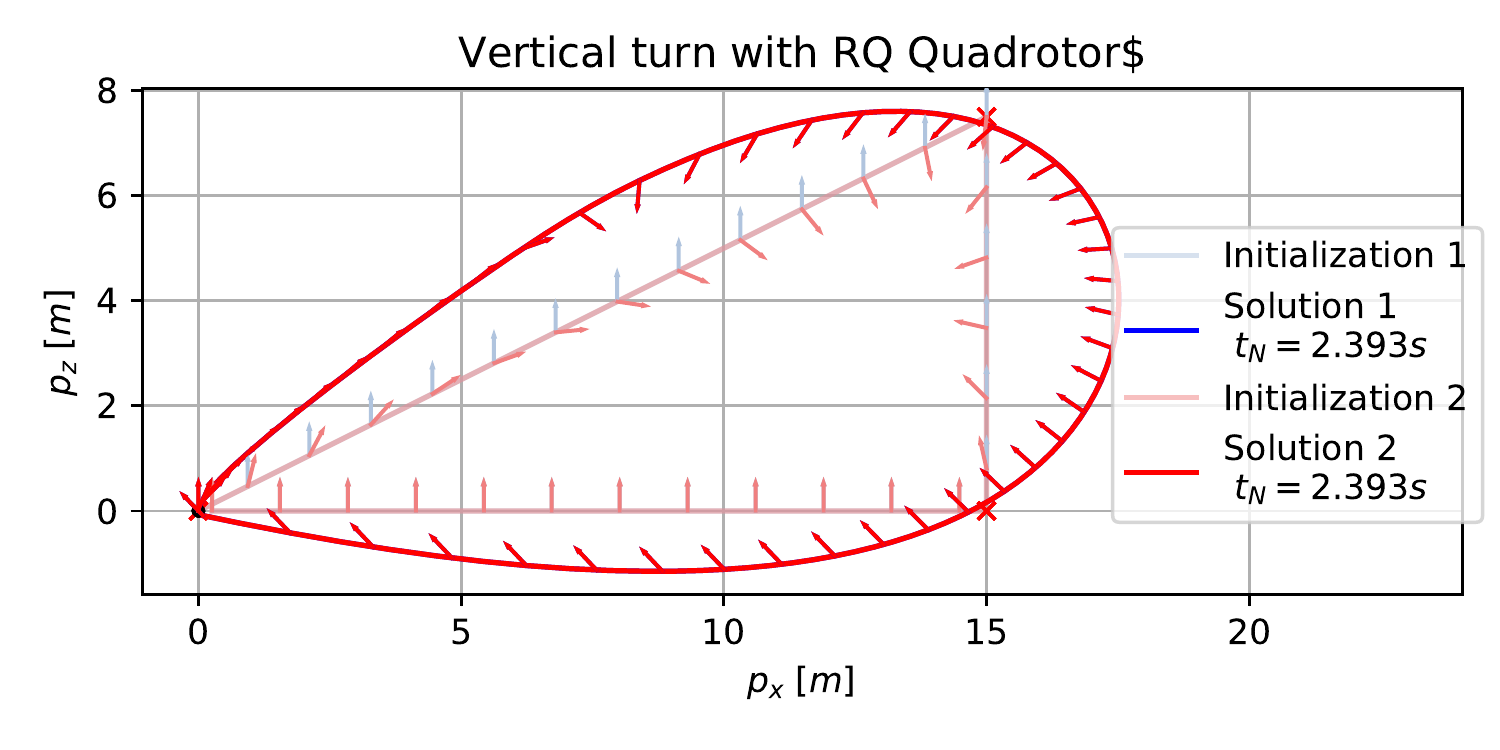}
    \end{subfigure}
    \caption{A vertical turn flown starting at the origin in hover and passing through both waypoints from top to bottom, and back to the origin.
    Two quadrotor configurations are used (STD in Fig. \ref{fig:exp_vertpin_std} and RQ in Fig. \ref{fig:exp_vertpin_rq}), with two different initialization setups each.
    The first setup is as described in \ref{sec:experiments} with identity orientation, while the second setup uses a linearly interpolated orientation guess.
    The arrows indicate the thrust direction of the quadrotor.
    The STD quadrotor configuration converges to two different solutions in Fig. \ref{fig:exp_vertpin_std} depending on the initialization due to its non-convex acceleration space with $T_{min} > \SI{0}{\newton}$, while the RQ quadrotor configuration converges to equal (and overlaying) solutions in Fig. \ref{fig:exp_vertpin_rq}, due to its convex acceleration space with $T_{min} = \SI{0}{\newton}$.
    }
    \label{fig:exp_vertpin}
\end{figure*}

\subsection{Initialization \& Convergence}
\label{sec:exp_initconv}
As a next step, the method is tested for convergence properties given different initializations.
For this, we again use the STD quadrotor configuration and discretize the problem into $N=160$ nodes with a tolerance of $d_{tol} = \SI{0.4}{\meter}$.
We use a track consisting of a so-called open hairpin, consisting of two waypoints as seen on the $xy$-plane in Fig. \ref{fig:exp_hairpin}, starting on the top left and passing the waypoint to the far right and bottom left, where the end point is not in hover.
Two setups are tested, where the first one is initialized with the position interpolated between the waypoints as in Fig. \ref{fig:exp_hairpin_pos}, and the second one is initialized with a poor guess interpolated only from start to end point, as in Fig. \ref{fig:exp_hairpin_init_pos}.

The expected outcome is that both initialization setups should converge to the same solution.
Indeed we can observe this behavior in Fig. \ref{fig:exp_hairpin}, where we depict the initial position guess in red and the convergence from light to dark blue.
The good initial guess in Fig. \ref{fig:exp_hairpin_pos} results in $216$ iterations until convergence, while the poor guess in Fig. \ref{fig:exp_hairpin_init_pos} needs $303$ iterations, all plotted in Fig. \ref{fig:exp_hairpin}.
This indicates that our method is not sensitive to correct initializations, but profits from good guesses.
However, since the acceleration space of a quadrotor is not necessarily convex and the resulting problem is highly non-convex, the next experiment elaborates on how to provoke and circumvent this issue. 

\subsection{Provoking Non-Convexity Issues}
\label{sec:exp_nonconv}
Since quadrotors can only produce thrust in body $z$-axis and the motors of most real-world systems cannot turn off, and therefore always produce a positive minimal thrust $T_{min} > 0$, the resulting acceleration space is non-convex.
We evaluate this on a vertical turn where we fly from hover at the origin through two waypoints directly above each other, back to the origin but not in hover.
The Track can be seen in Fig. \ref{fig:exp_vertpin}.
First, the STD quadrotor configuration is used, with $N=150$ nodes and a tolerance of $d_{tol}=\SI{0.1}{\meter}$, with the general initialization setup where the orientation is kept at identity.
A second setup uses a different initialization, where we interpolate the orientation around the $y$-axis between $\alpha_{init} = 0$, $\alpha_0 = \pi$ for the second waypoint and $\alpha_1 = \alpha_2 = 2 \pi$ for the remaining waypoints.
The solutions and associated initializations are depicted in Fig. \ref{fig:exp_vertpin_std}, in which it is obvious that they do not converge to the same solution.
The second setup actually performs a flip which is slightly faster at $\SI{4.115}{\second}$ compared to the first setup at $\SI{4.171}{\second}$ ($1.3\%$ faster).
This is expected due to the non-convex properties of the problem.

However, a second set of experiments is performed with the same initialization setups but the RQ quadrotor configuration.
This configuration has a minimum thrust of $T_{min} = \SI{0}{\newton}$, which renders the achievable acceleration space convex.
Both setups for the RQ configuration are depicted in Fig.\ref{fig:exp_vertpin_rq}.
Indeed, both initializations now converge to the same solution, which overlay each other and achieve the same timing at $\SI{2.393}{\second}$.

A simple solution would be to first solve the same problem using a linear and therefore convex point-mass model and using this to initialize the problem with the full quadrotor model.
We further elaborate on the convexity property in the discussion Sec. \ref{sec:dis_convexity}.

\subsection{Slalom}
\label{sec:exp_slalom}
\begin{figure}[t]
\caption*{Slalom Track \quad $t_N = \SI{8.644}{\second}$}
\begin{subfigure}{\linewidth}
    \centering
    \caption{$xy$-Position}
    \label{fig:exp_slalom_pos}
    \includegraphics[width=\linewidth, trim={0 0 0 6}, clip]{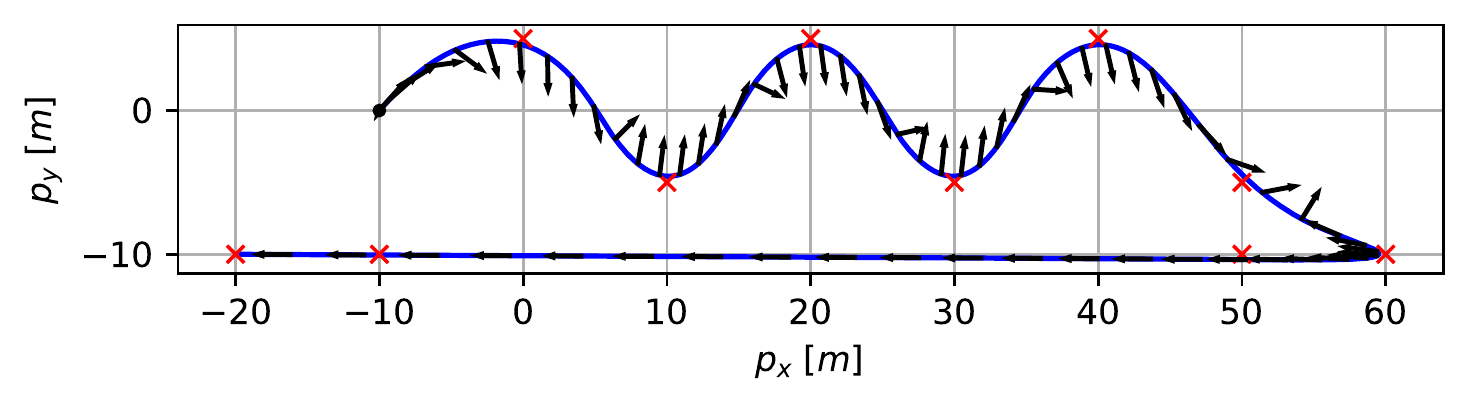}
\end{subfigure}
\begin{subfigure}{\linewidth}
    \centering
    \caption{Velocity Profile}
    \label{fig:exp_slalom_vel}
    \includegraphics[width=\linewidth, trim={0 0 0 6}, clip]{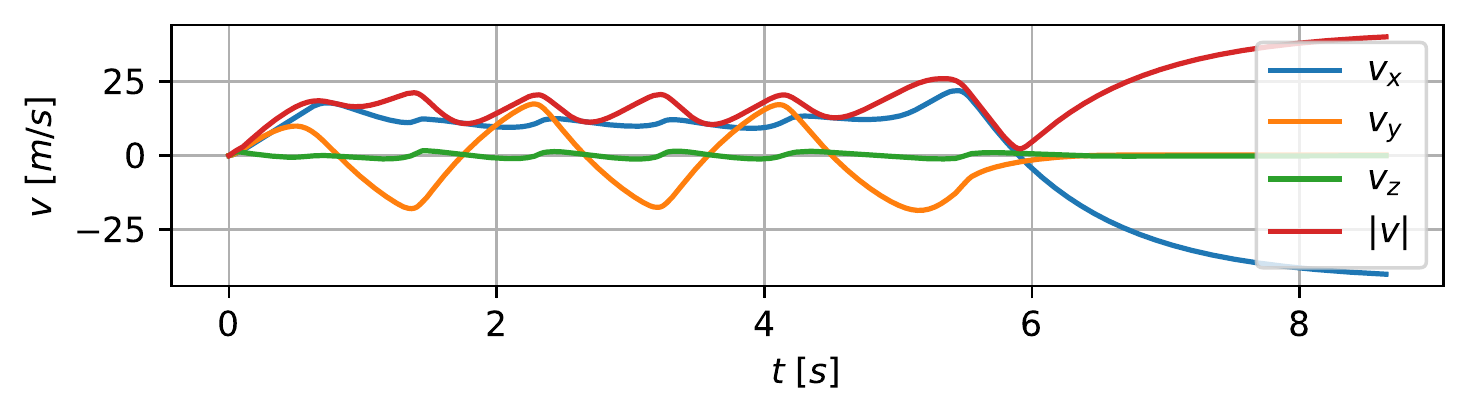}
\end{subfigure}
\begin{subfigure}{\linewidth}
    \centering
    \caption{Progress Variables}
    \label{fig:exp_slalom_prog}
    \includegraphics[width=\linewidth, trim={0 0 0 6}, clip]{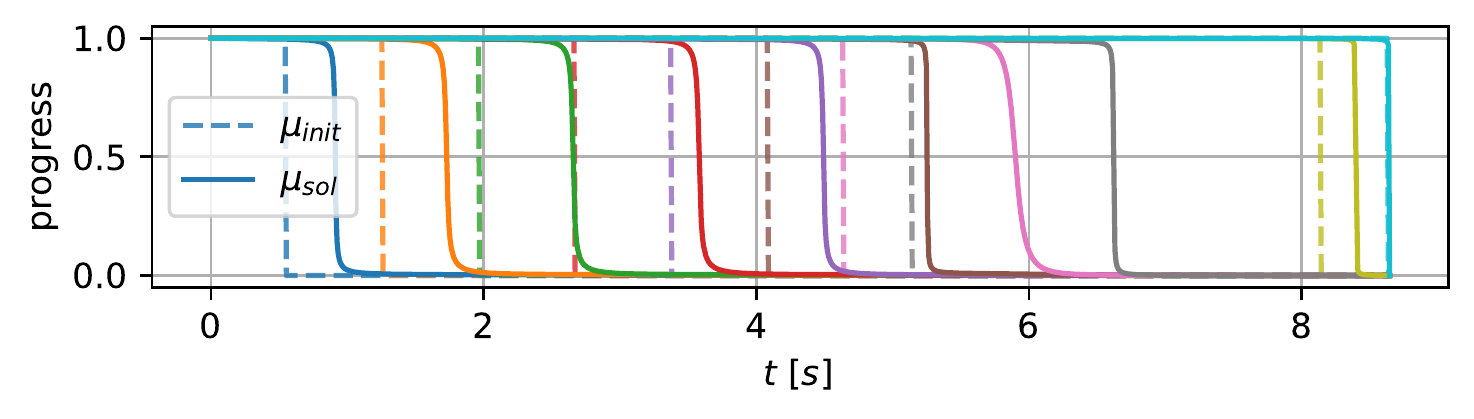}
\end{subfigure}
\caption{A track consisting of a slalom, an open hairpin and a long stretch back, flown with the RQ quadrotor configuration in $t_N = \SI{8.644}{\second}$.
In Fig. \ref{fig:exp_slalom_pos} the black arrows denote the acceleration direction.
Note how the velocity approaches a steady state due to the linear aerodynamic drag on this quadrotor, and how the progress variables are being significantly shifted from their initial guess.}
\label{fig:exp_slalom}
\end{figure}

In a next step, a longer track is created, including a slalom and a long straight part, with 10 waypoints in total, as in Fig. \ref{fig:exp_slalom}.
The trajectory is discretized into $N=800$ nodes with a tolerance of $d_{tol}=\SI{0.4}{\meter}$.
The RC quadrotor configuration is deployed in this experiment.
We expect the trajectory to smoothly slalom through the first 6 waypoints, take a sharp turn and accelerate back through the long straight segment.
The straight segment is on purpose spanned by 4 unequally spaced waypoints, which together with the 6 previous waypoint provide a poor initial guess for the progress variables.

The resulting time-optimal trajectory is plotted in $xy$-plane in Fig. \ref{fig:exp_slalom_pos}, with the velocity in Fig. \ref{fig:exp_slalom_vel} and the progress variables in Fig. \ref{fig:exp_slalom_prog}.
Note the dashed initial guess for the progress variables in Fig. \ref{fig:exp_slalom_prog}, which are significantly shifted to enable a time-optimal waypoint allocation.
Furthermore, the velocity approaches its limit at the end of the straight segment, due to linear aerodynamic drag in this quadrotor configuration.

\subsection{Microsoft AirSim, NeurIPS 2019 Qualification 1}
\label{sec:exp_airsim}
As an additional demonstration, we apply our algorithm on a track from the 2019 NeurIPS AirSim Drone racing Challenge \cite{Madaan20arxiv}, specifically on the Qualifier Tier 1 setup.
We choose a quadrotor with roughly the same properties, described as the MS configuration in Tab. \ref{tab:quads}.
The track is set up with the initial pose and 21 waypoints as defined in the environment provided in \cite{Madaan20arxiv}\footnote{\url{https://github.com/microsoft/AirSim-NeurIPS2019-Drone-Racing}, accessed in May 2020}.
We use a discretization of $N = 3360$ nodes and a tolerance of $d_{tol} = \SI{0.1}{\meter}$.

The original work by \cite{Madaan20arxiv} also provides a simple and conservative baseline performance of $t_{total} \approx \SI{110}{\second}$ under maximal velocity and acceleration of $v_{max} = \SI{30}{\meter\per\second}$ and $a_{max} = \SI{15}{\meter\per\second\square}$, respectively.
However, the best team achieved a time of $t_{total} = \SI{30.11}{\second}$ according to the evaluation page\footnote{\url{https://microsoft.github.io/AirSim-NeurIPS2019-Drone-Racing/leaderboard.html}, accessed in May 2020}.
Our method generates a trajectory that passes all waypoints at a mere $t_N = \SI{24.11}{\second}$.
Please note that this trajectory should only serve as a theoretical lower bound on the possibly achievable time given the model parameters.

\begin{figure}[t]
\caption*{NeurIPS Airsim Qualification 1 Track \quad $t_N = \SI{24.11}{\second}$}
\vspace{3pt}
\begin{subfigure}{0.49\linewidth}
    \centering
    \caption{$xy$-Position}
    \label{fig:exp_airsim_q1_pos_xy}
    \vspace{-5pt}
    \includegraphics[width=\linewidth,trim={0 0 0 10},clip]{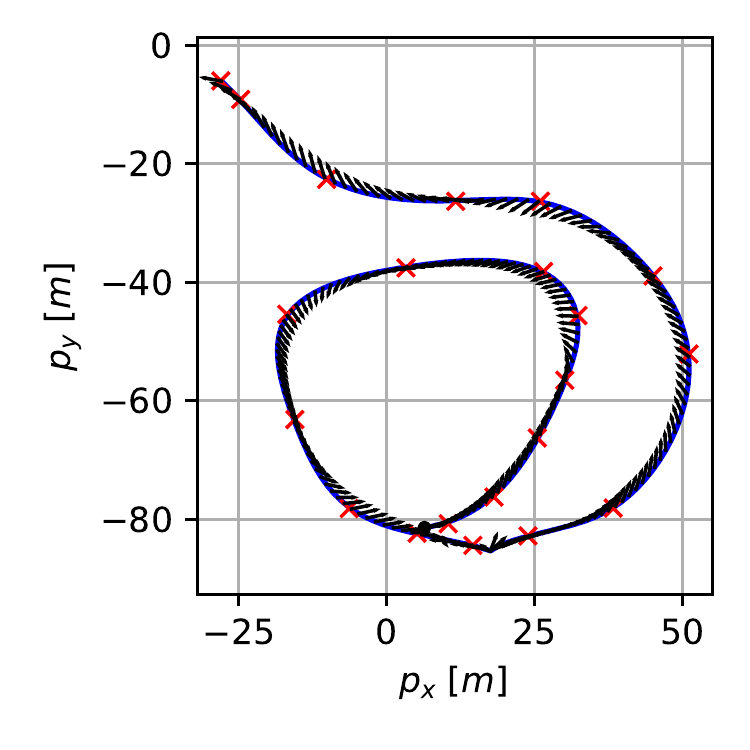}
\end{subfigure}
\hfill
\begin{subfigure}{0.49\linewidth}
    \centering
    \caption{$xz$-Position}
    \label{fig:exp_airsim_q1_pos_xz}
    \vspace{-5pt}
    \includegraphics[width=\linewidth,trim={0 0 0 10},clip]{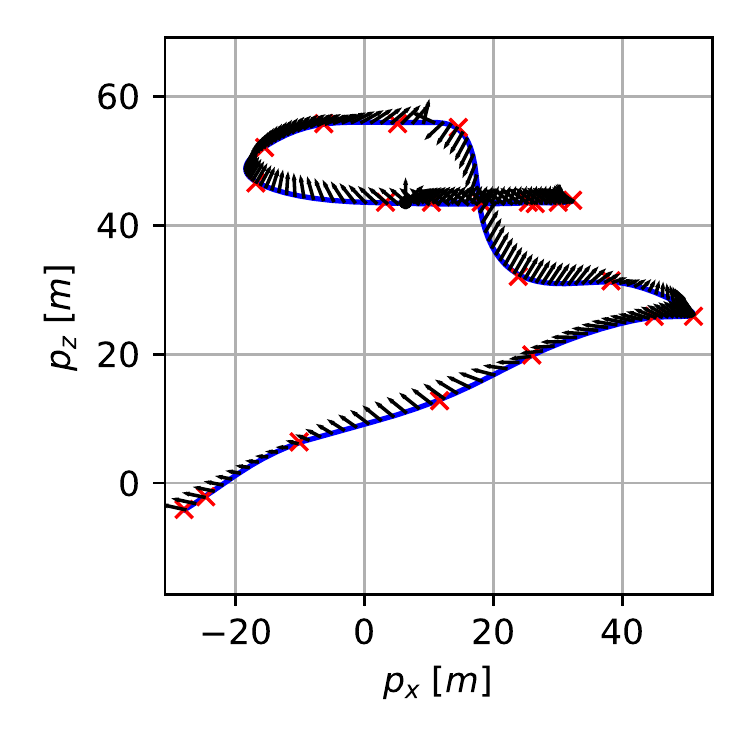}
\end{subfigure}\\
\begin{subfigure}{\linewidth}
    \centering
    \caption{Velocity Profile}
    \label{fig:exp_airism_q1_vel}
    \includegraphics[width=\linewidth,trim={0 0 0 10},clip]{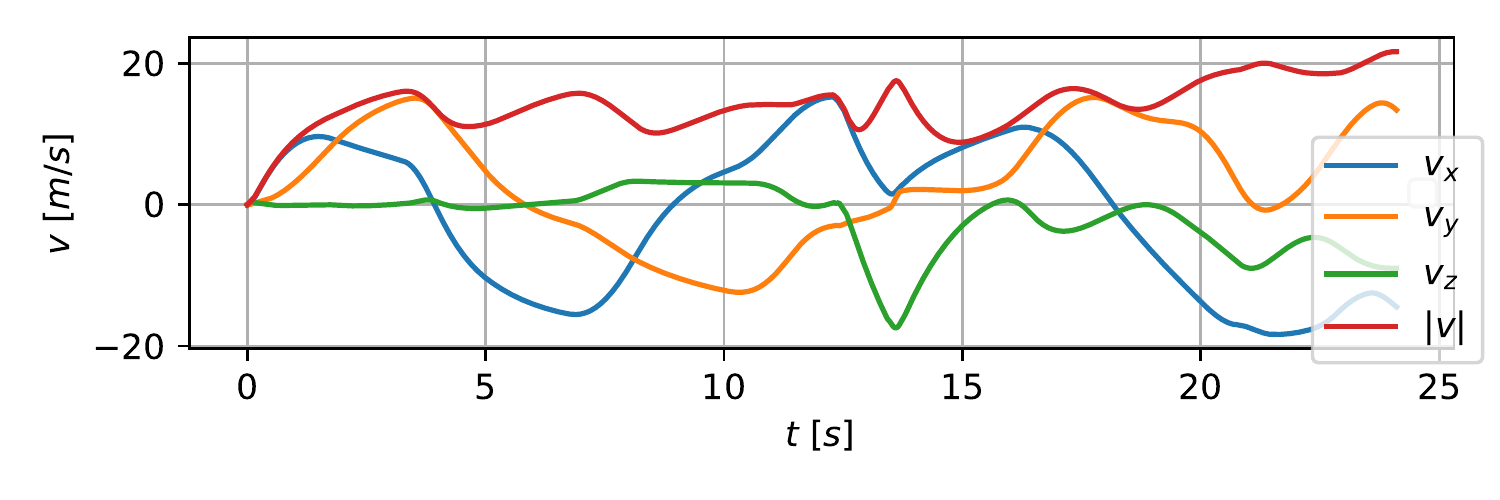}
\end{subfigure}
\begin{subfigure}{\linewidth}
    \centering
    \caption{Progress Variables}
    \label{fig:exp_airism_q1_prog}
    \includegraphics[width=\linewidth,trim={0 0 0 10},clip]{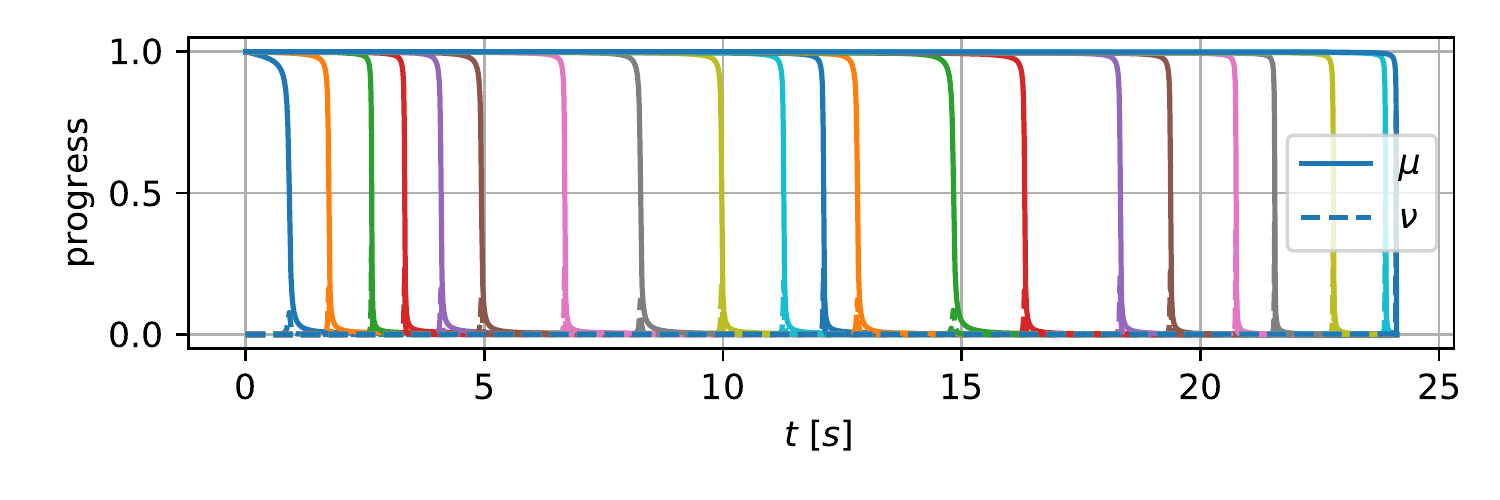}
\end{subfigure}
    \caption{The NeurIPS Airsim Qualification 1 track, covered in $t_N = \SI{24.11}{\second}$, as opposed to the best team's $\SI{30.11}{\second}$.
    The top row depicts the trajectory in $xy$- and $xy$-plane, while the second and thrid row depict the velocity and progress respectively, plotted over time.}
    \label{fig:exp_airsim_q1}
\end{figure}

\clearpage
\subsection{Qualitative Human-Comparison in Simulation}
\label{sec:exp_humansim}
In this experiment, a comparison to human-flown trajectories is established.
As a short foreword, we want to point out that the human trajectories were collected in an experimental third-party simulation, focused on good flight characteristics for human pilots, but enabling logging of the drone state.

\begin{figure}[]
    \centering
    \caption*{Comparison to Human Flight in Simulation}
    \includegraphics[width=\linewidth,trim={12 10 0 25},clip]{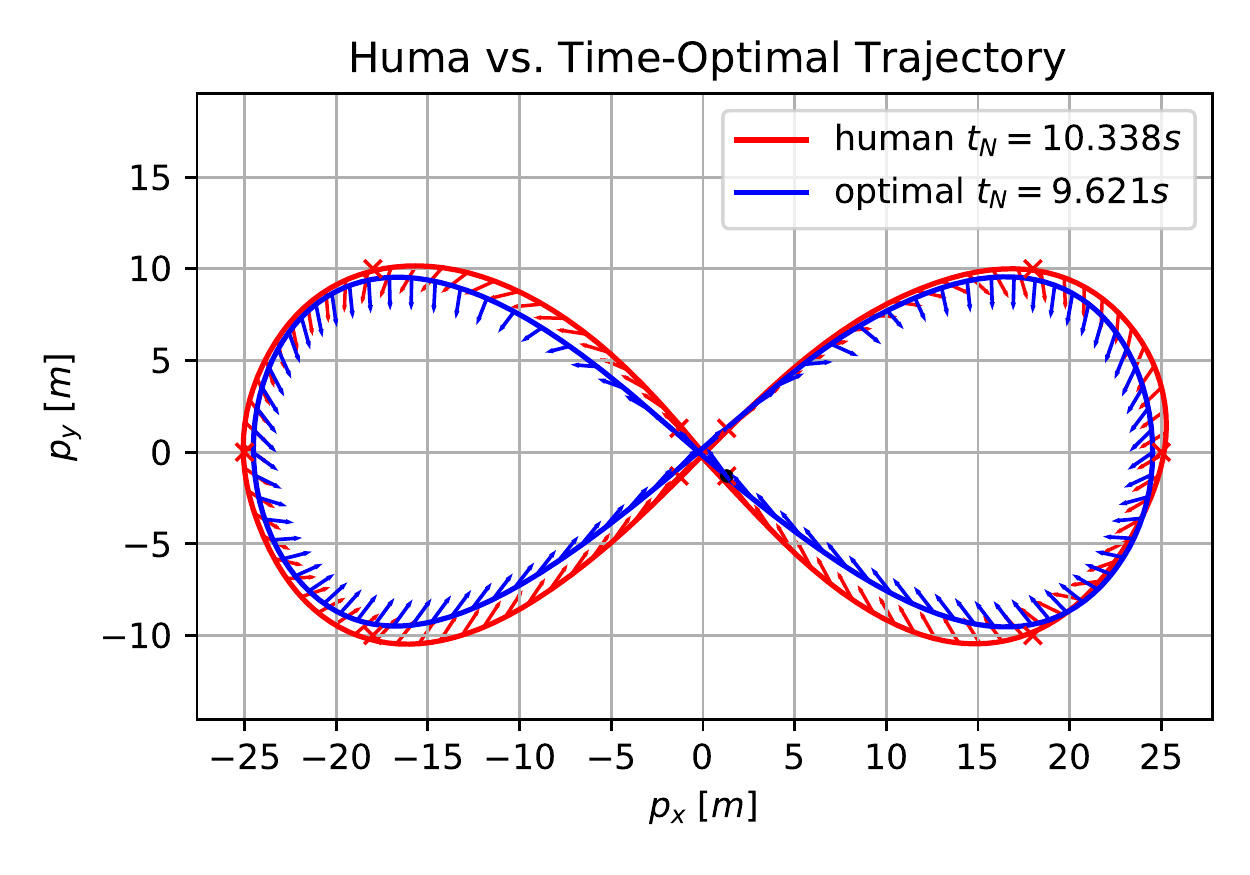}
    \caption{Comparison of a human trajectory of $t_N = \SI{10.338}{\second}$ flown in simulation by an expert pilot and a time optimal trajectory of $t_N = \SI{9.621}{\second}$ generated by our proposed method.
    Note that the human pilot is commanding collective thrust and bodyrates, which are tracked by a low-level controller.
    This means that the human can not fully exploit the single rotor thrust and therefore the true actuation limit.}
    \label{fig:exp_sim}
\end{figure}

We setup the experiment as a figure-8-shaped track as depicted in Fig. \ref{fig:exp_sim}.
The used quadrotor model is the SIM configuration listed in Tab. \ref{tab:quads}.
While the mass, inertia, size, and drag constant is set as in the nominal model, the maximum thrust and bodyrates were fixed to the maximum values the human pilot was able to achieve with this platform in the simulation for a more fair comparison. (corresponding to the listed values in Tab. \ref{tab:quads}).
Therefore, the maximum bodyrate was set to $\omega_{max} = \SI{3}{\radian\per\second}$, and the maximum thrust to $T_{max} = \SI{16}{\newton}$.
We use $N = 1000$ discretization nodes and a tolerance of $d_{tol} = \SI{0.4}{\meter}$.

The human had multiple tries for preparation and also at testing time $>10$ consecutive rounds through the figure-8 track were collected, where one round starts and ends at the trajectory point passing closest to the origin, corresponding to the center intersection.
The best lap with the lowest time  was used and is depicted in Fig. \ref{fig:exp_sim}, with a time of $\SI{10.338}{\second}$.
The generated time-optimal trajectory takes only $\SI{9.621}{\second}$, exploiting the maximum available thrust at all times, except for when introducing and stopping quick rotations to realign the thrust through the gate passes.

Note that the human commands bodyrate and collective thrust, where the single rotor thrust control is left to a simulated low-level controller, as usual for human drone flight.
This however imposes limits on how much of the input space the human can exploit, which is especially significant given the low thrust-to-weight ratio of this drone configuration.
Due to these limitations, we also perform a real-world comparison of a high-speed maneuver in the next section.

\subsection{Qualitative Human-Comparison in Real-World}
\label{sec:exp_humanreal}
\begin{figure}
    \caption*{Comparison to Real-World Human Flight}
    \begin{subfigure}{0.49\linewidth}
    \caption{$xy$-Position}
    \label{fig:exp_real_hairpin_pos_xy}
    \vspace{-6pt}
    \includegraphics[width=\textwidth,trim={5 0 0 0},clip]{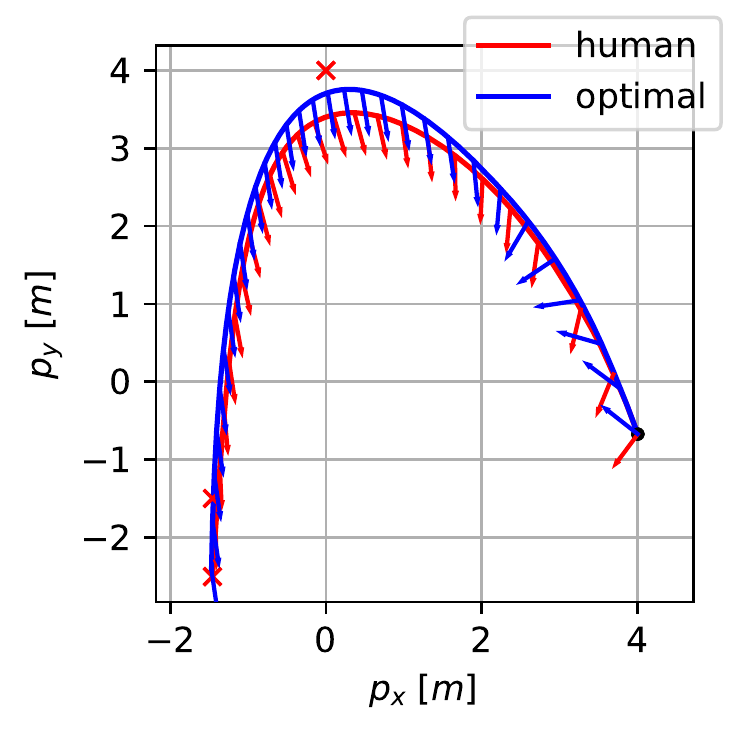}
    \end{subfigure}
    \hfill
    \begin{subfigure}{0.49\linewidth}
    \caption{Acceleration Space}
    \label{fig:exp_real_hairpin_acc}
    \includegraphics[width=\textwidth,trim={0 0 5 20},clip]{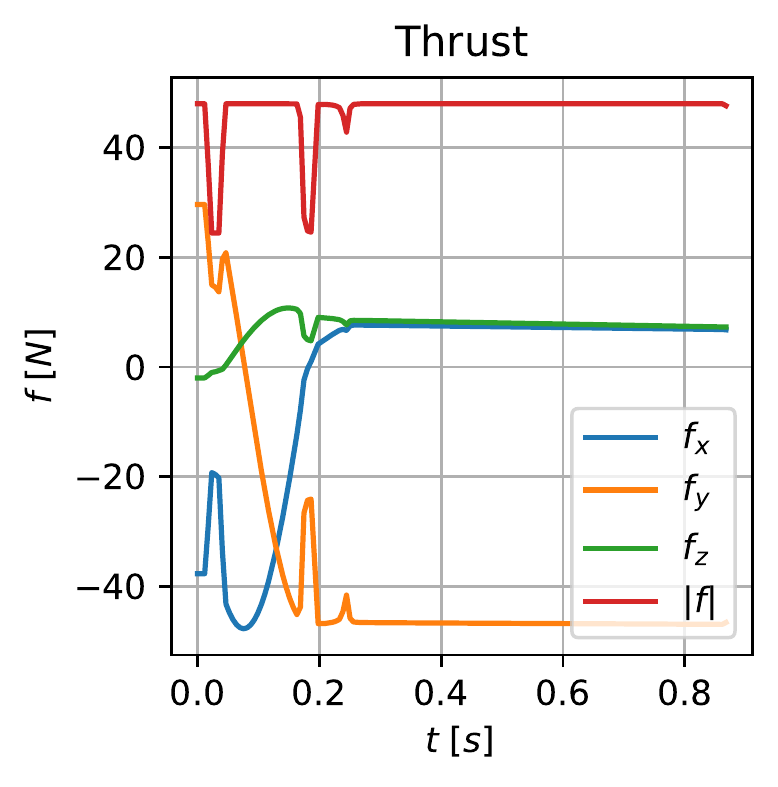}
    \end{subfigure}
    \caption{Comparison of a human-flown real-world trajectory to a time-optimal trajectory, starting in the lower right corner (see Fig. \ref{fig:exp_real_hairpin_pos_xy}).
    The human-flown maneuver was recorded in a motion capture system and timed at $t_{human} = \SI{0.984}{\second}$ while the optimal trajectory reached a timing of $t_N = \SI{0.874}{\second}$.
    The optimal trajectory exploits the full acceleration of the quadrotor as long as possible (see Fig. \ref{fig:exp_real_hairpin_acc}), and rotates into deceleration later than the human (see Fig. \ref{fig:exp_real_hairpin_pos_xy}).}
    \label{fig:exp_real_hairpin}
\end{figure}

As a last experiment, we provide a qualitative comparison of a time-optimal trajectory to a real human-flown trajectory recorded in a motion capture system.
The trajectory represents a hairpin turn around a pole (see Fig. \ref{fig:exp_real_hairpin}), which was picked because of its simplicity and repeatability for the human, and because it is a common element in drone racing where often the maximum available acceleration is exploited.
the optimization is initialized at the point where the human pilot entered the trackable region and initial position and velocity of the vehicle are set equal to the human's, while the orientation and bodyrate is left open for optimization.
We then add 3 waypoints forcing the drone to pass around the pole further away than the human, and leading the trajectory to exit at the same point and with roughly the same exit direction as the human.
The problem is solved with $N=150$ nodes and $d_{tol} = \SI{0.4}{\meter}$ using the RC quadrotor configuration, modelled after the human's real quadrotor used for this experiment.
As in the previous experiment, we restrict the maximum thrust to the same limit the human was able to exploit over a significant period of time as measured by the onboard IMU.

The maneuver is extremely quick and even the human trajectory only lasts for $\SI{0.984}{\second}$, while the optimal solution reduces this to $\SI{0.874}{\second}$.
We can see that the time-optimal solution has a different acceleration direction at the beginning (see Fig. \ref{fig:exp_real_hairpin_pos_xy}), and starts to break later than the human trajectory, achieving a slightly better time.
Additionally, we plot the thrust (absolute and directional) in Fig. \ref{fig:exp_real_hairpin_acc} and see the common behavior of the optimization exploiting the maximum thrust, only lowering it to adjust the rotational rates at the beginning.

Please note that, as before, the optimization uses the nominal model and is not aware of complex aerodynamic effects, noise or disturbances.
It should serve as a theoretical lower-bound on time, guidance for what could be achieved, and as a demonstrator that our method does indeed work.

\section{Discussion}
\label{sec:discussion}

\subsection{Convexity}
\label{sec:dis_convexity}
While the problem of trajectory optimization quickly becomes non-convex when using complex and/or non-linear dynamic models, constraints, or even cost formulations (e.g. obstacle avoidance), it is often a valid approach to generate feasible (in terms of model dynamics) and near optimal trajectories.
In our experiments, we have provoked and demonstrated one such non-convex property, and also quickly elaborate how one could support the optimization with an advanced initialization scheme to start close to the global optimum.

This can be achieved by reducing the non-linear quadrotor model into a linear point-mass model with bounded 3D acceleration input $\bm{u} = \bm{a}$ where $\|\bm{a}\| \leq a_{max}$.
This linear model renders the problem convex and allows to find a solution from which the original problem with the quadrotor model can be initialized, both in terms of translational trajectory, and also with a non-continuous orientation guess based on the point-mass acceleration direction.
While this initial guess is not yet a dynamically feasible trajectory due to the absent rotational dynamics, it serves as a good initial guess in close proximity to the optimal solution.

\subsection{Optimality}
\label{sec:dis_optimality}
Our approach does not provide optimality guarantees, since the problem by definition is non-convex and the optimal solution, while always guaranteed to be locally optimal, does not necessarily need to be globally optimal.
Additionally, the used implementation framework \cite{Andersson18} and solver backend \cite{Waechter06jmp} might influence the convergence behavior and the applicable guarantees.
The previously described initialization scheme allows to mitigate these problems and, by running multiple setups (e.g. with different orientation guesses), one could cross check different solutions to find the optimal one with certainty (comparable to sampling methods).

\subsection{Real-World Applicability}
\label{sec:dis_applicability}
There are two problems hindering our approach from being applied in real world scenarios.

The first problem is posed by the nature of time-optimal trajectories themselves, as the true solution for a given platform is nearly always at the actuator constraints, and leaves no control authority.
This means that even the smallest disturbance could potentially have fatal consequences for the drone and render the remainder of the trajectory unreachable.
One has to define a margin lowering the actuator constraints used for the trajectory generation to add control authority and therefore robustness against disturbances.
However, this also leads to a slower solution, which is no longer the platform-specific time-optimal one.
In the context of a competition, this effectively becomes a risk-management problem with interesting connections to game theory.

Second, our method is computationally demanding, on a modern laptop taking many minutes ($\approx 1-40 \si{\minute}$) for scenarios as in Sec. \ref{sec:exp_p2p}-\ref{sec:exp_nonconv} or sometimes even hours for larger scenarios such as \ref{sec:exp_slalom}-\ref{sec:exp_humansim}.
However, this is highly implementation-dependent and could be vastly broken down to usable times, or precomputed for static race tracks and other non-dynamic environments.

\section{Conclusion}
In this work, we proposed a novel complementary progress constraint that allows to dynamically allocate waypoints to unknown nodes on a trajectory, and therefore find a time-optimal waypoint-flight solution without having to sample or allocate waypoint-completion timings.
We validated our method with many experiments in a bottom-up fashion, checking against existing work, verifying the dynamic time allocation, provoking and elaborating on unwanted non-convex properties and initialization dependence.
Finally, we demonstrated our method on longer tracks with more than $M \geq 10$ waypoints, and in a qualitative comparison against a human expert.
We conclude that our method can be used to generate theoretical time-optimal trajectories at the actuation limit of the platform, which 
serve as an upper-bound of the achievable performance and guidance to develop fast and agile quadrotor flight.
\section{Acknowledgements}
The authors want to thank Elia Kaufmann, Thomas Laengle, Christian Pfeiffer, and our professional pilot and collaborator Gabriel Kocher for their support and contribution throughout this work. 
This work was supported by the National Centre of Competence in Research Robotics (NCCR) through the Swiss National Science Foundation, the SNSF-ERC Starting Grant, and the EU H2020 Research and Innovation Program through the AERIAL-CORE project (H2020-2019-871479).

\balance
\bibliographystyle{unsrtnat}
\bibliography{references}

\end{document}